\documentclass[review,onecolumn, 1p,times,authoryear]{elsarticle}

\usepackage{amssymb}
\usepackage{amsmath}
\usepackage{amsthm}

\usepackage{tabularx}

\usepackage{bbm} 
\usepackage{cancel} 
\usepackage{layouts} 
\usepackage{adjustbox} 
\usepackage{rotating} 
\usepackage{svg} 
\usepackage{caption} 
\usepackage{subcaption} 
\usepackage{tabularx}
 \usepackage{booktabs}
\usepackage{multirow}

\newtheoremstyle{mystyle}
{10pt}
{10pt}
{\itshape}
{}
{\bfseries}
{}
{0.5em}
{\thmname{#1}\thmnumber{ #2}:\thmnote{ #3}}

\theoremstyle{mystyle}

\theoremstyle{definition}

\newtheorem{rquestion}{Research Question (RQ)}

\newcommand{\myparagraph}[1]{\paragraph{}\mbox{}\\}

\usepackage{moreverb,url}
\usepackage[bookmarksopen,bookmarksnumbered]{hyperref}

\usepackage{cleveref}
\creflabelformat{equation}{#2\textup{#1}#3} 
\usepackage[nolist]{acronym}

\journal{Journal of Decision Systems}

\begin{document}

\begin{acronym}
\acro{EE}[EE]{evaluation episode}
\acro{DSR}[DSR]{design science research}
\acro{RQ}[RQ]{research question}
\acro{RBF}[RBF]{radial basis function}
\acro{MMD}[MMD]{maximum mean discrepancy}
\acro{ML}[ML]{machine learning}
\acro{CS}[CS]{computer science}
\acro{IS}[IS]{information systems}
\acro{EM}[EM]{expectation maximization}
\acro{PAC}[PAC]{probably approximately correct}
\acro{MLA}[MLA]{maximum likelihood approach}
\acro{GAN}[GAN]{generative adversarial network}

\end{acronym}

\begin{frontmatter}



\author{Philipp Spitzer\fnref{kit, equal}}
\author{Dominik Martin\corref{cor1}\fnref{kit,tss, equal}}
\author{Laurin Eichberger\fnref{kit}}
\author{Niklas Kühl\fnref{ubt}}


\title{Towards a Problem-Oriented Domain Adaptation Framework for Machine Learning}


\affiliation[kit]{organization={Karlsruhe Institute of Technology},
            addressline={Kaiserstr. 89},
            city={Karlsruhe},
            postcode={76133},
            country={Germany}}
\affiliation[tss]{organization={Trelleborg Sealing Solutions},
            addressline={Schockenriedstr. 1},
            city={Stuttgart},
            postcode={70565},
            country={Germany}}
\affiliation[ubt]{organization={University of Bayreuth},
            addressline={Wittelsbacherring 10},
            city={Bayreuth},
            postcode={70565},
            country={Germany}}
\fntext[equal]{These authors contributed equally to this work.}





\begin{abstract}
Domain adaptation is a sub-field of machine learning that involves transferring knowledge from a source domain to perform the same task in the target domain. It is a typical challenge in machine learning that arises, e.g., when data is obtained from various sources or when using a data basis that changes over time.
Recent advances in the field offer promising methods, but it is still challenging for researchers and practitioners to determine if domain adaptation is suitable for a given problem---and, subsequently, to select the appropriate approach.
This article employs design science research to develop a problem-oriented framework for domain adaptation, which is matured in three evaluation episodes. We describe a framework that distinguishes between five domain adaptation scenarios, provides recommendations for addressing each scenario, and offers guidelines for determining if a problem falls into one of these scenarios. During the multiple evaluation episodes, the framework is tested on artificial and real-world datasets and an experimental study involving 100 participants. 
The evaluation demonstrates that the framework has the explanatory power to capture any domain adaptation problem effectively. In summary, we provide clear guidance for researchers and practitioners who want to employ domain adaptation but lack in-depth knowledge of the possibilities.  
\end{abstract}




\end{frontmatter}


\section{Introduction}
\label{sec:introduction}

In \ac{ML}, domain adaptation refers to approaches that bridge the gap between two problems that share the same task but differ in the feature distribution \citep{zhuang2020}. Domain adaptation aims to leverage knowledge about one problem (i.e., the source domain) to find a solution for the other problem (i.e., the target domain). Transferring knowledge from one domain to another is often desired in practical applications of supervised learning, e.g., if labeled data in the target domain can hardly be collected or is more expensive than in the source domain \citep{kouw2019}.

In autonomous driving, for instance, reliable object recognition models to navigate the self-driving car and detect pedestrians, other cars, traffic signs, etc. are required. Typically, however, there is only a limited number of annotated pictures of road scenes available to train a model in a supervised manner, as manually labeling these pictures is expensive. Instead, synthetic data, i.e., rendered traffic scenes, are easy to generate and, thus, cheap \citep{cordts2016}. However, usually, models tend to perform well on these synthetic data, but the performance drops significantly when applied to real-world images caused by a domain shift, i.e., the synthetic pictures do not exactly behave or look like the ones from the real-world \citep{hoffman2018}.

Even if researchers or \ac{ML} practitioners suspect to profit from domain adaptation for a given problem, there is little actionable guidance on selecting a promising approach from the ever-growing pool of domain adaptation techniques and implementations. Thus, we propose a \emph{problem-oriented domain adaptation framework} that enables users to identify and classify domain adaptation problems, delivers guidance on suitable solution approaches, and helps to pinpoint reasons for ill-performing domain adaptation attempts. The framework is developed and presented according to the \ac{DSR} paradigm as described by \citep{hevner2004} to ensure a user-centered design of the artifact. Furthermore, by drawing on the \ac{DSR} paradigm, we acknowledge the importance of demonstrating practical implications in real-world contexts while ensuring theoretical rigor. This alignment with \ac{DSR} allows us to effectively bridge the gap between theory and practice, making our framework valuable for both researchers and practitioners dealing with domain adaptation scenarios \citep{hevner2007}.


\subsection{Theoretical and Practical Relevance} 
\label{sec:introduction:relevance}

Expanding from the motivational examples at the beginning, in this section, we highlight the importance of domain adaptation in \ac{CS} and \ac{IS} as well as the relevance of actionable decision support in the application domains.

We highlight the \emph{theoretical relevance} by connecting current state-of-the-art theories and algorithms in \ac{CS} to the application of domain adaptation, which is a special case of transfer learning. Transfer learning is an important aspect of \ac{ML} because it enables the application of \ac{ML} methods for difficult problems by sharing knowledge between different but similar problem settings. If such an approach yields better results than handling each problem independently, we speak of positive transfer (effects) \citep{pan2010}. Different problem settings often show different feature dimensions or different feature distributions. This is called a domain shift \citep{wang2018}. One reason for non-positive transfer between problem settings is that a domain shift violates the fundamental independent identical distribution (i.i.d.) that is required in statistical learning during the minimization of the expected error \citep{hutchison2010, pan2010, yang2007}. Domain adaptation represents transfer learning approaches that deal with domain shifts and enable positive transfer between one problem scenario to another, even if the underlying feature distributions differ \citep{weiss2016}. 

\begin{figure}
    \centering
    \includegraphics[width=0.9\linewidth]{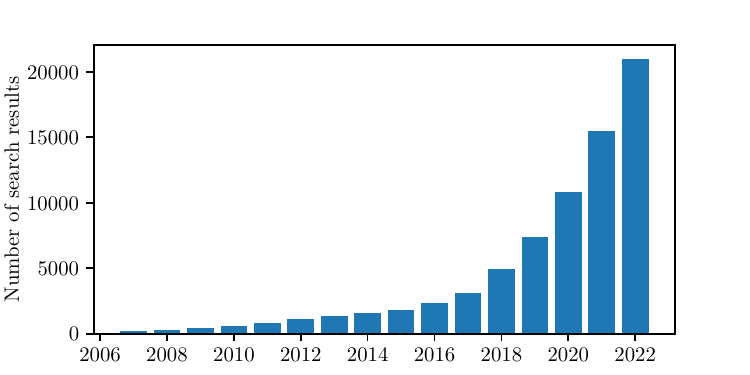}
    \caption[Search results for domain adaptation by year.]{The number of search results for \emph{domain adaptation} in Google Scholar by year indicates a growing interest in the topic.}
    \label{fig:SearchResults}
\end{figure}

In recent years, the research concerning domain adaptation has significantly increased, as can be seen in the upwards trend of search results for the term presented in \Cref{fig:SearchResults}. This rise is partly due to advances in the theoretical foundations of domain adaptation, e.g., limits for the upper error bound concerning the learning error \citep{ben-david2007, zhao2019}. Likewise, techniques from other fields of \ac{ML} are transferred successfully into the domain adaptation context. Prominent examples are deep learning for domain adaptation (e.g., CAN  \citep{kang2019}), the utilization of adversarial architectures for domain mapping (e.g., CyCADA \citep{hoffman2018}), or domain invariant feature learning (e.g., DANN \citep{ajakan2015}).

Still, some authors disagree on the viability of many theoretical approaches in real-life scenarios \citep{zhao2019, wilson2020, kouw2019}.
Thus, the domain adaptation framework presented in this work tries to bridge theory and practice by condensing and simplifying the complexity of state-of-the-art research that can be overwhelming even for domain experts. To our knowledge, such a framework focusing on what is relevant in practical scenarios and prioritizing streamlined implementation over cutting-edge but often highly specialized and theoretical algorithms does not yet exist.

Continuing from how domain adaptation as a transfer learning mechanism is relevant in \ac{ML} theory, the \emph{practical relevance} depends on the ability of researchers of different fields to use domain adaptation in their work and create novel insights based on the achieved results. To assess the contribution of the proposed framework, we emphasize the practical relevance of domain adaptation in real-world applications and motivate how a framework can support the described use cases. 

Supervised learning is a type of \ac{ML} which makes up 95\% of all \ac{ML} problems \citep{Jordan2015}. In supervised \ac{ML}, a model is trained to predict a target label given input by learning from labeled examples \citep{russell2010}. As this approach requires high quantities of such labeled examples, referred to as training data, it runs into problems where labeled training data is not readily available. For many practical use cases, there is no or only very few labeled training data available, which in turn limits the usefulness of supervised learning approaches. 

\sloppy Creating labeled examples (e.g., through manually labeling data) is often expensive or not even possible to perform on the required scale. For example, in the well-known Cityscapes dataset, each pixel-level annotated image took on average 1.5 hours to complete \citep{cordts2016}. Obviously, this fact calls for alternatives. Besides semi-supervised \citep{zhu2007, zhuang2020} or unsupervised learning \citep{hinton1999} which inherently require only a few or no labels at all, intuitively, another possibility is to look for other but similar training data, where labels are readily available, and train a model on this data. This approach of transferring the knowledge from this related model to the original problem so that an increase in performance for the original problem is observable is called transfer learning \citep{pan2010}. If in a transfer learning scenario, the actual task of the model remains the same throughout all problem settings (e.g., recognizing digits from visual input). Consequently, the problems differ only in their domain (i.e., the feature space and the marginal distribution of the features). For example, if one domain consists of pictures of handwritten digits from 0 to 9 whereas another domain contains images of digits taken from photos of house numbers, it is a reasonable assumption that while these domains are somehow similar, a model that is trained on only the handwritten digits might fail to deliver the same results if presented with the house number digits. 

\begin{figure}
    \centering
    \includegraphics[width=\linewidth]{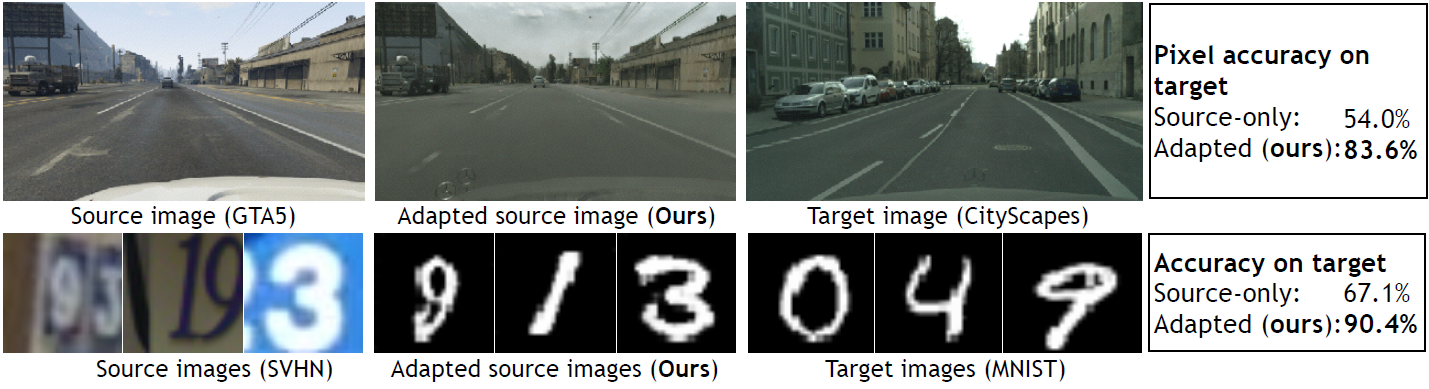}
    \caption[Domain adaptation results for CyCada.]{Domain adaptation results for CyCada by \citet{hoffman2018}. The adversarial approach aligns the source domain (GTA5, SVHN) to more closely resemble the target domain (CityScapes, MNIST) and achieves higher accuracy in the respective computer vision tasks compared to only learning from the source domain.}
    \label{fig:CyCada}
\end{figure}

A positive transfer of knowledge, therefore, requires that the difference in the domains between the problem settings is somehow bridged. Domain adaptation is a term that describes approaches designed for overcoming this so-called domain divergence \citep{wang2018}. \Cref{fig:CyCada} presents an adversarial domain adaptation approach by \citet{hoffman2018} that deals exactly with this challenge. 

Domain adaptation is a central aspect of transfer learning if the task remains the same for all scenarios. Apart from the motivating example of mitigating the need for high quantities of labeled data in the target domain, domain adaptation can also be used to integrate data from different sources. For example, integrating records and observations of different teams that use differing measuring equipment was performed for medical scans made in different hospitals \citep{kamnitsas2016, opbroek2015}. Another common cause for a domain divergence is the problem of the sample selection bias, which is relevant to many research settings in statistics and econometrics \citep{lee1982}. It is therefore no surprise that examples for applications of domain adaptation can be found in almost every application field of \ac{ML}. This includes computer vision \citep{Zhang2018, chen2018}, natural language processing \citep{preotiuc-pietro2017, zhang2017}, forecasting \citep{hosseini-asl2018, shinohara2016} and various forms of domain generalization \citep{muandet2013, zhao2017, ghifary2015}.

\begin{figure}
    \centering
    \includegraphics[width=0.9\linewidth]{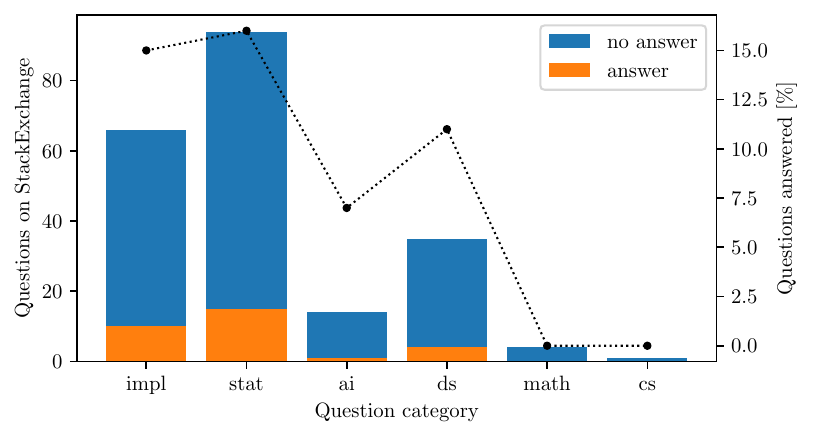}
    \caption[Answer rates on the StackExchange network.]{Answer rates for questions concerning domain adaptation on the StackExchange network are significantly below the average of 70\%. The questions are categorized by the topics: implementation (impl), statistics (stat), artificial intelligence (ai), data science (ds), mathematics (math), and computer science (cs). The data is obtained using the official API as of April 2023.}
    \label{fig:AnswerRatesOnTheStackExchangeNetwork}
\end{figure}

The large variety of applications demonstrates the practical relevance of domain adaptation in \ac{ML} but does not necessarily imply the need for a practical framework for domain adaptation as presented in this work. We acknowledge, that it is difficult to find examples in scientific literature that refer to implementation issues or concerns. However, this could be due to the fact that scientific articles are positively-descriptive by nature. This means that often results are only published if they are considered a success. Thus, it is unlikely that a literature review yields valuable information on authors' difficulties in applying domain adaptation in research projects and is even less likely to find information on if domain adaptation due to practical challenges was not considered or even failed. 
To bridge this information gap on challenges in practice, we draw on publicly available information on issues. The web page \emph{stackoverflow.com} is a popular question-and-answer site for programmers with over 100 million monthly views in April 2023 \citep{stackoverflow2023}. It belongs to a network of similarly structured websites for different subjects, such as statistics, \ac{CS}, etc. called the StackExchange.  As of April 2023, the StackExchange network has over 40 million users \citep{stackexchange2023}. By querying the StackExchange network for questions concerning domain adaptation, it becomes apparent that a majority of questions remain unanswered. The results are illustrated in \Cref{fig:AnswerRatesOnTheStackExchangeNetwork}. On average about 70\% of questions are answered for all topics compared with only circa 15\% for domain adaptation. The low answer rate can be considered an indicator of a lack of practical support. 

\subsection{Research Objectives and Scope} 
\label{sec:introduction:scope}

Domain adaptation serves as a powerful \ac{ML} technique for utilizing data from a similar yet distinct domain, where data is more readily accessible than in the original domain, to solve a task in the original domain. The efficacy of domain adaptation is highly dependent on domain characteristics, and the successful application of domain adaptation methods often requires a deep understanding of both the domain and the underlying theory \citep{wilson2020}. As reasoned in \Cref{sec:introduction:relevance}, it is reasonable to assume that researchers and practitioners may be hesitant to employ domain adaptation due to the resources required for evaluating its potential benefits and subsequently applying it to their projects. This leads to the central research questions:

\begin{rquestion}
\label{rq:1}
\emph{How can the appropriateness of domain adaptation for a specific problem be determined a priori using practical indicators?}
\end{rquestion}

\begin{rquestion}
\label{rq:2}
\emph{How can domain adaptation problems be meaningfully distinguished to suggest targeted solution strategies?}
\end{rquestion}

To address these questions, we propose an artifact in the form of a conceptual framework with the primary objective of assisting users in meaningfully performing domain adaptation. From the overall objective, we derive three major design requirements: (1) the framework specifies diverse, generally applicable, and selectively addressable domain adaptation scenarios, (2) practitioners can a priori and with reasonable effort determine the correct scenario for any given problem, and (3) effective solution approaches tailored for each domain adaptation scenario are provided.

To focus on the core deliverable, the following scope is defined: The framework is limited to single-source, homogeneous, unsupervised domain adaptation problems. Intuitively, this implies that we only consider two domains (one source and one target domain) sharing the same feature space (e.g., both domains contain gray-scale 24x24 pixel images), and no target labels are available. The exact definitions of the terms used above are presented in \Ref{sec:appendix:concepts:tl}. This limitation is common in domain adaptation as it represents the base case problem. In-depth empirical evaluations of domain adaptation algorithms are not carried out in this work, instead, relevant related literature providing comprehensive insights is referenced.

\subsection{Contributions}\label{sec:introduction:contributions}
To summarize the contributions of this work, the proposed framework significantly advances the field of domain adaptation in machine learning by delivering a novel, problem-oriented domain adaptation perspective. They can be summarized as follows:
\begin{enumerate} 
    \item Development of a Comprehensive Framework: We introduce a structured framework that categorizes domain adaptation into five distinct scenarios, providing clear and actionable guidelines to assist in determining the suitability of domain adaptation for specific problems, a feature notably lacking in the existing literature. Thus, we support practitioners and researchers in the field alike.
    \item Empirical Validation: The proposed framework has been rigorously evaluated across multiple datasets and an experimental study involving 100 participants, demonstrating its robustness and versatility in various domain adaptation contexts.
    \item Bridging Theory and Practice: Our approach is grounded in rigorous theoretical research while also being keenly attuned to practical applications, effectively bridging the gap between theoretical concepts and real-world applicability. By providing a clear categorization and decision-making process, our framework makes the complex field of domain adaptation more accessible to a broader audience, including those with limited background in the area.
\end{enumerate}

These contributions collectively represent a significant step forward in the domain adaptation domain, offering both theoretical insights and practical tools for researchers and practitioners.

\section{Related Work}
\label{sec:related_work}
Domain adaptation is a sub-field of transfer learning. \citet{zhuang2020} published a survey on transfer learning, which discusses numerous approaches and serves as a starting point. This survey builds on previous surveys by \citet{weiss2016, pan2010} on the same topic. \citet{zhang2020} provide another survey on transfer learning, focusing on the historical development of solution approaches. More specific surveys on domain adaptation include single-source unsupervised cases \citep{kouw2019} and deep domain adaptation techniques \citep{wang2018, wilson2020}. Additionally, \citet{wilson2020} offer the most comprehensive overview of empirical results for various deep domain adaptation approaches and problems. 

Moreover, there are plenty of articles presenting different approaches to domain adaptation, of which deep domain adaptation (i.e., domain adaptation using deep neural networks) is the most prominent \citep{wu2022imbalanced, liu2024cross, jimenez2021study}, for instance, to cope with distribution shifts of electrocardiograms \cite{he2023novel}. However, this short summary does not aim to be a comprehensive overview of methods present in recent literature. The articles listed here should be understood as representatives for a multitude of implementations. 
The idea of using adversarial domain adaptation gained traction after the proposal of ``Domain-Adversarial Training of Neural Networks'' by \citet{ganin2016domain}. Because adversarial training yielded promising results for many domain adaptation tasks, various modern approaches picked up the idea. Examples of this are Co-DA \citep{kumar2018}, CyCADA \citep{hoffman2018} (cf. \Cref{fig:CyCada}) or DIRT-T \citep{shu2018}. The authors in \cite{hong2024robust} improve the performance of entropy-based adversarial networks by supplementing the discriminator with additional probability values. Their results show superior performance to state-of-the-art methods. In a similar work of \cite{lee2023domain}, the authors present an approach to tackle the simultaneous occurrence of domain shift and class imbalance based on adversarial networks. Through various evaluations, they show that their method, based on learning domain-invariant features and applying a label-aligned sampling strategy, improves the performance compared to conventional methods. Nevertheless, there are also deep domain adaptation approaches that are not based on adversarial learning, e.g., \citet{french2018} use self-ensembling which has delivered at least state-of-the-art performance for some popular scenarios. The study \cite{taghiyarrenani2023multi} presents a method to cope with distribution shifts in regression tasks. Their method builds on Siamese neural networks and shows its effectiveness in multi-domain settings. Similarly, \cite{liu2022end} outline how to deal with domain shifts in remote sensing domains. Their approach improves the performance of state-of-the-art models in cross-domain change detection. Next to deep domain adaptation frameworks, there are approaches that do not rely on deep neural networks. Instances of these are class-based reweighting \citep{saerens2002} or sample-based reweighting \citep{huang2006} for correcting sample selection bias for unlabeled data.

Further articles explore domain adaptation by examining the statistical properties of domain shifts more closely. \citet{kouw2019a} employ the same definitions for classifying dataset shifts based on \citet{moreno-torres2012} as we do in this work. However, their application of these principles is less rigorous, as they do not place the same emphasis on the causality of the underlying problem. In addition, \citet{zhang2013} provides another example of applying dataset shift definitions in domain adaptation.

Recent works provide a comprehensive set of methods to deal with domain shifts in various settings. However, the question of when to apply domain adaptation is highly relevant for the successful use and implementation of these methods in real-life settings. Few works provide guidance for the successful application of domain adaptation methods \cite{kouw2019, kouw2019a, ben-david2007, zhao2019, bashath2022data, chang2020systematic}. The work of \cite{chang2020systematic} informs about the use of unsupervised domain adaptation methods for wearable sensors. In their work, the authors provide guidelines for practitioners on how to apply domain adaptation approaches in this domain. Furthermore, both works by Kouw et al. explore when to apply domain adaptation from a theoretical perspective \citep{kouw2019a, kouw2019}. Error bounds are another way to define the suitability of certain domain adaptation approaches. Bounds for domain adaptation are presented by \cite{ben-david2007} or \cite{zhao2019}, of which the latter is significantly more tailored towards modern domain adaptation. In the work of \cite{bashath2022data}, the authors present a data-centric perspective of deep learning models in text data domains. They provide a new nomenclature and taxonomy of deep learning approaches for the application of domain adaptations with text data. 

Looking at domain adaptation from an implementation perspective, there is a framework called ``Semi-supervised Adaptive Learning Across Domains'', in short SALAD, that provides tools to perform experiments for some domain adaptation approaches \citep{schneider2018}. It was last updated in 2018. Besides a pre-release of the same authors for domain adaptation toolbox that is especially tailored towards ImageNet \citep{2021a}, we could not find any further studies that are targeted towards \ac{ML} practitioners and are sufficiently broadly applicable.

With the vast development of domain adaptation methods in various domains and few works providing theoretical or practical details on the application of such methods, guidelines for applying domain adaptation a priori in specific contexts are missing. Thus, in this work, we develop a framework to facilitate the use in research and practice when to apply domain adaptation and give tailored solution recommendations.

\section{Methodology}
\label{sec:methodology}
The underlying research paradigm which is used for this work is \ac{DSR}. \ac{DSR} is an outcome-based methodology that focuses on generating knowledge via the design or improvement of usable artifacts. It was first developed by \citet{simon1996} and later refined by among others \citet{hevner2010} and \citet{vanaken2005}. \citet{hevner2010} describe \ac{DSR} as ``a research paradigm in which a designer answers questions relevant to human problems via the creation of innovative artifacts, thereby contributing new knowledge to the body of scientific evidence'' \citep[p. 5]{hevner2010}.

At its core, this classifies \ac{DSR} as constructive research that can be differentiated from explanatory research. In explanatory research, practical implications are often derived as a last step based on theoretical foundations. The decoupling of practice and theory meant that research can fail to provide any significant results for practical applications. A problem that was often encountered during the early days of \ac{CS} \citep{iivari2005}. As a result, researchers and practitioners developed a new approach that focuses on putting the product or the solution to an actual problem first and then derives knowledge and novel theories from the construction of the product, which in \ac{DSR} terms is referred to as an \emph{artifact}. Putting the product first is the essence of constructive research and has shown to be a valid strategy for tackling complicated problem settings.

The general design cycle was first analyzed by \citet{takeda1990} and later adapted by \citet{vaishnavi2007} to specifically target \ac{DSR}. The design phase of the artifact follows the notion of the design cycle and the \ac{DSR} methodology as proposed by \citet{peffers2007}. Both approaches require (intermediate and final) evaluation to guide the development and measure the suitability of the final artifact. A framework for constructing such \acp{EE} in \ac{DSR} is proposed by \citet{venable2016}. Its implementation for this article is explained in \Cref{sec:evaluation}. The general intuition is that formative artificial evaluation is used throughout the design to guide the development and more naturalistic and summative evaluation scenarios are used to test the result of the development.

\begin{figure*}
    \centering
    \includegraphics[width=0.83\textwidth]{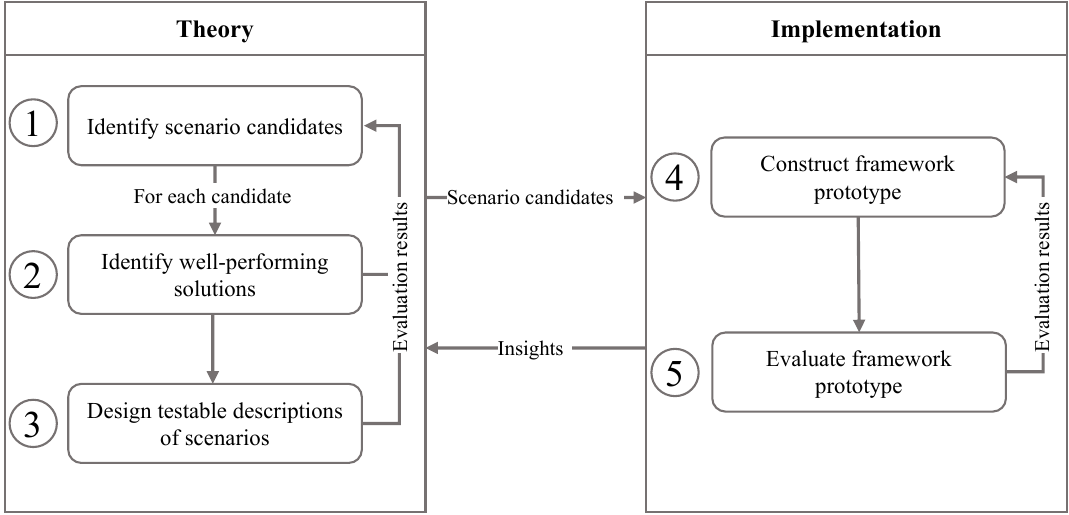}
    \caption[Artifact design road map.]{The road map illustrates how the artifact, i.e., the framework, is constructed and improved by the two interconnected steps: theory building and implementing/evaluating. The process can be repeated if finished to incorporate insights from a previous iteration. Alternatively, intermediary results can lead to an immediate revisiting of earlier steps.}
    \label{fig:research_process}
\end{figure*}

Given the exploratory nature of our research, we have adopted an iterative design approach, linking intermediate findings and including technical evaluations to validate the design. \Cref{fig:research_process} delineates our research methodology, which encompasses five iterative steps: 1) identifying potential domain adaptation scenarios from empirical evidence and theory, 2) selecting effective algorithms or types for each identified scenario, 3) creating testable descriptions for these scenarios, 4) developing a framework that coherently integrates scenarios with solution strategies based on solid theoretical grounds, and 5) evaluating this framework through both descriptive and observational methods.

\section{A Problem-Oriented Domain Adaptation Framework}
\label{sec:artifact}

The core artifact is the identification of four (respectively five) domain adaptation scenarios that are the result of applying Moreno-Torres' unified view on data shifts \citep{moreno-torres2012} in the context of a \ac{ML} problem, more precisely an unsupervised homogeneous domain adaptation problem as previously defined. This is not a completely new approach. For instance, in their survey of domain adaptation methodologies, \citet{kouw2019a} acknowledge the perspective of \citet{moreno-torres2012} regarding data shifts, although it is not strictly incorporated. Many sources do, if at all, only superficially mention the connection and focus much more on specific details in the respective domain adaptation approach. An example of this is ``Domain Adaptation Under Target And Conditional Shift'' by \citet{zhang2013}. During the research, it became more evident that in order to provide qualified support for practitioners, one cannot rely on a set of statistical markers, e.g., domain divergence via Jenson-Shannon divergence, alone, but must understand the characteristics of domain adaptation problems on a more basic level. This is because actual measures are confronted with a wide range of practical problems, e.g., the curse of dimensionality \citep{bellman1957dynamic}, the difficulties in estimating probability distributions in regions with sparse sample coverage \citep{silverman1986density}, and extreme variations in the empirical divergence measures based on entropy for regions in which source and target differ significantly \citep{lee2000algorithms}. As a result, a central question of the framework is: \emph{what is the causality of the system?}
This question is almost always overlooked in domain adaptation literature \citep{pearl2009}. It is not only a core component of Moreno-Torres' view on data shift, but also highly important in interpreting what we know about a given domain adaptation problem correctly in order to choose a suitable solution approach as will be shown later in this section.

The concept of \emph{causality} can be briefly explained as follows: Anyone familiar with \ac{ML} expects a problem where a label or value $y$ is predicted based on observable features $x$ with the help of experience that is learned from samples. Often the samples are said to be created by an unknown system with joint probability $P(x,y)$. In the real world, there is a connection present between $x$ and $y$ that is defined by the causal relationship. If $X$ is the reason for $Y$, the notation is $X \rightarrow Y$. An example of this is predicting the immediate failure probability $y$ of an engine based on various sensor readings $x$. The alternative is a system where $Y$ is the reason for $X$, i.e., $Y \rightarrow X$. This is the case, if we try, e.g., to recognize animals $y$ on image data $x$. Depending on the causality of the system, the joint probability $P(x \cap y)$ can be decomposed into  

\begin{itemize}
    \item $P(x\cap y)=P(y|x)\cdot P(x)$ for $X \to Y$ systems, or
    \item $P(x\cap y)=P(x|y)\cdot P(y)$ for $Y \to X$ systems.
\end{itemize}
The important fact is, that although we always predict values of $Y$ based on observations of $X$, the causal reality can be the other way around. 

\Cref{tab:scenarios} on page \pageref{tab:scenarios} shows the basic domain adaptation scenarios (prior shift, covariate shift, concept shift, class-conditional shift, and data set shift) in accordance with the definitions by \citet{moreno-torres2012} and how the probabilities change for each scenario. To help the reader differentiate between inferred and given properties, the properties given by the definition are written in bold letters. In the following sub-chapters, each scenario and the corresponding domain adaptation procedures will be introduced in detail. The suggested solution procedures for each scenario are summarized in \Cref{tab:SolutionProcedures} and support for determining the correct scenario is given in \Cref{tab:ScenarioDetermination}.

\begin{table*}
\caption[Domain adaptation scenarios]{The domain adaptation scenarios are based on the shift types identified by \citet{moreno-torres2012}. Each column starting with $\Delta$ represents a shift in the corresponding probability distribution. Shifts that are prescribed by the definition are written in bold letters.}
\label{tab:scenarios}
\centering
\small
\begin{tabular}{l l *{5}{c}}
\toprule
Causality           &  Shift type        & $\Delta P(y)$  & $\Delta P(x)$ & $\Delta P(x|y)$  & $\Delta P(y|x)$  & $\Delta P(x \cap y)$                         \\ \hline
\multirow{2}{*}{$Y \rightarrow X$} & Prior      & \textbf{yes} & yes &\textbf{no}  & yes & yes \\
                    & Class-cond.   & \textbf{no}  & yes & \textbf{yes} & yes & yes                     \\ \hline
\multirow{2}{*}{$X \rightarrow Y$} & Covariate  & yes & \textbf{yes} & yes & \textbf{no}  & yes          \\
                    & Concept    & yes & \textbf{no}  & yes & \textbf{yes} & yes                    \\ \hline
Unspecified             & General    & yes & yes & yes & yes & yes                   \\ 
\bottomrule
\end{tabular}
\end{table*}

\begin{table*}
\caption[Solution procedures summary]{Summary of solution procedures}
    \label{tab:SolutionProcedures}
    \centering
    \small
    \begin{tabularx}{\linewidth}{l l X}
    \toprule
    \multicolumn{2}{l}{Scenario} & Solution procedure  \\ \cmidrule{1-2}
         Causality& Shift& \\
         \midrule
         $ Y \to X$ & Prior & Perform class-based reweighting, either by  (a) design, if the accuracy is not equally important for all classes (e.g., disease detection), or (b) algorithm choice to align the prior distributions and minimize the overall risk, e.g., by \ac{EM}-algorithm \citep{saerens2002}. 
         
         See also: cost-sensitive learning \citep{elkan2001}, learning from imbalanced data \citep{he2009}.
\\
& Class-cond. & Approximate the relationship between the domains by defining a transformation function $t$ so that $P_s(x)=P_t(t(x))$, e.g., by (a) subspace mapping, (b) finding domain invariant components or (c) feature augmentation. For many problems, this is usually done via \acp{GAN}, e.g., Co-DA \citep{kumar2018}. 

Further information: Surveys by \citet{wilson2020} or \citet{wang2018}. \\
\midrule
$X \to Y$ & Covariate & Perform instance-based reweighting to align the posterior distributions to minimize the error if the classifier is misspecified. Otherwise, train on the highest number of available samples. 

See also: sample selection bias \citep{huang2006, lee1982}.\\
& Concept & No domain adaptation scenario, because the task is changing explicitly without changes in the feature distribution. 

See also: concept drift \citep{karimian2023concept, widmer1996, gama2014, zliobaite2010}.  \\
\midrule
Unspecified & General & Depending on the causality, apply solution procedures of class-cond. or concept shift. The performance depends on the magnitude of the shifts and on the implementation details of the selected approach. There is no performance guarantee. To minimize sources of error, incorporate as much information about the domain relationship as possible. \\
    \bottomrule
    \end{tabularx}
\end{table*}

\subsection{Prior Shift}
\label{sec:artifact:prior_shift}
A prior shift, or prior probability shift, appears only in $Y \rightarrow X$ problems and is defined as the case where $P_s(x|y)=P_t(x|y)=P(x|y)$ and $P_s(y) \neq P_t(y)$ \citep{moreno-torres2012}. An example from \ac{ML} for a prior shift is a binary computer vision task, in which images of dogs and cats must be correctly classified and the source and target domain differ in the share of dog pictures, e.g., in the source domain we have $P_s(y=\text{dog})=P_s(y=\text{cat})=0.5$ and in the target domain $P_t(y=\text{dog})=0.7 \neq P_t(y=\text{cat})=0.3$. On the other hand, the class conditionals remain the same, i.e., in both domains, a picture of a dog looks the same. 

\begin{figure}
    \centering
    \includegraphics[width=\linewidth]{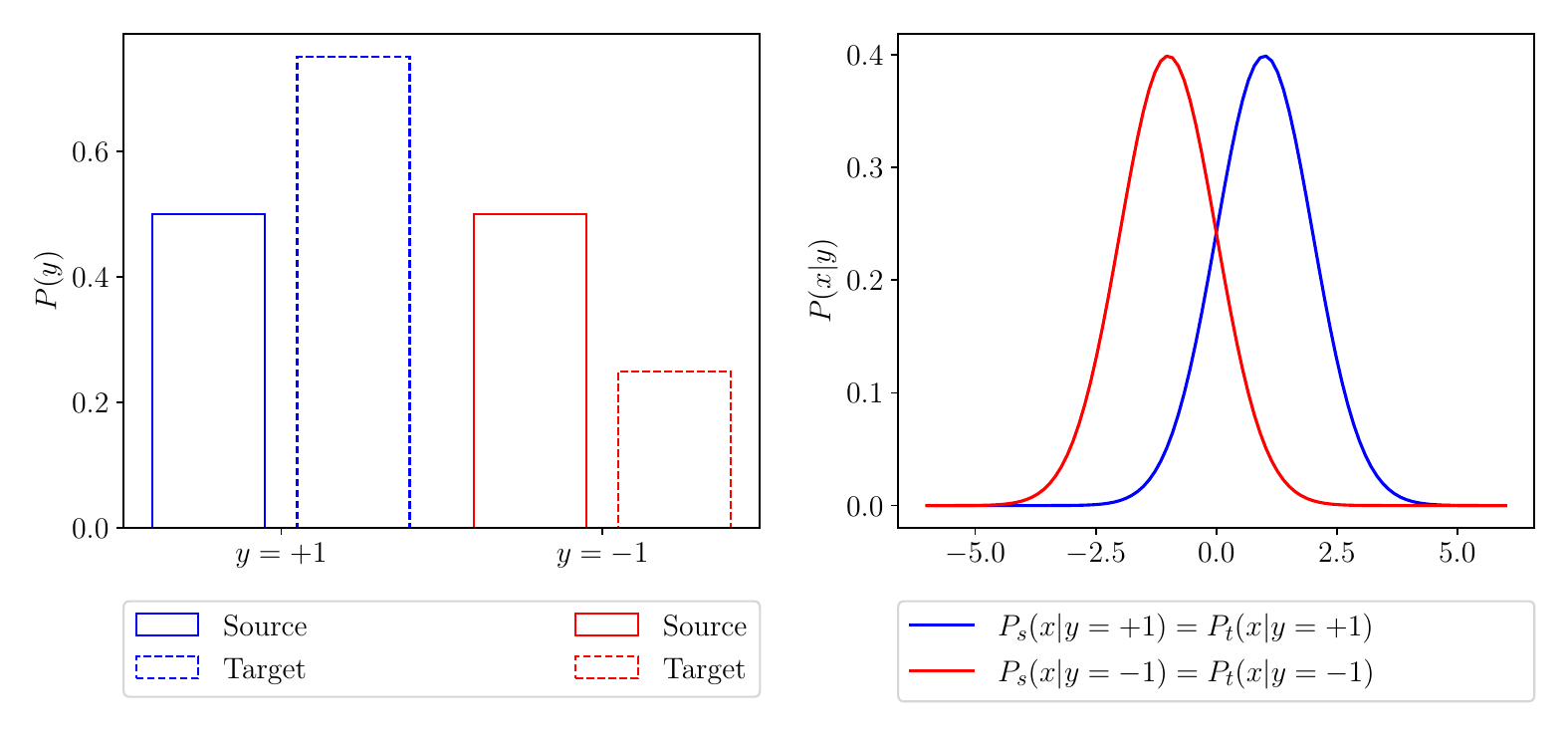}
    \caption[Prior shift distributions.]{In $Y \rightarrow X$ systems, a \textbf{prior shift} occurs if the marginal label distributions (left) change, but the class conditionals (right) remain the same.}
    \label{fig:PriorShift}
\end{figure}

This knowledge can be exploited in what is sometimes referred to as class-based reweighting \citep{he2009}. If the risk function on the target domain is expressed via a decomposition of the joint probability into prior and conditional, we see that the conditionals are canceled out, because they are equal in the source and target domain per definition. \citep{saerens2002}. 

\begin{equation}
\label{eq:RiskPriorShift}
R_t(h)=\sum_{y\in Y} \int_{\mathcal{X}}\ell\left(h(x),y \right) \frac{\cancel{P(x|y)}P_t(y)}{\cancel{P(x|y)}P_s(y)}P_s(x,y)dx    
\end{equation}

This means that an estimator of the target risk can be obtained using only source labels and a weighting function. 

\begin{equation}
\label{eq:EmpiricalRiskPriorShift}
\hat{R}_t(h)=\frac{1}{n}\sum^{n}_{i=1}\ell \left(h(x_i),y_i \right)w(y_i)
\end{equation}

Obtaining this weighting function without target labels is one domain adaptation approach that was used by researchers even before the term domain adaptation itself became popular. In literature, it is often referred to as class balancing (for unsupervised learning). A possible approach that uses an \ac{EM} approach is presented by \citet{saerens2002}, in which the samples of the target domain are soft labeled by the source classifier based on the maximum likelihood. This can be seen as an implementation of the ML learning framework (cf. \Cref{def:MlaLearning}). The authors have shown empirically that for low divergence prior shifts and poorly chosen samples, this approach is prone to actually negatively impact a classifier's performance and delivered a statistical test to decide if a reweighting is beneficial. \citet{zhang2013} present a reweighting method that is based on quadratic optimization for a kernel mean matching approach to avoid the direct estimation of probability distributions and observe positive effects on the model performance \citep{zhang2013}. 
Correcting a prior shift is also an objective in cost-sensitive learning \citep{elkan2001} and learning from imbalanced data \citep{he2009}, which has been studied outside of the domain adaptation context. 

When deciding to correct for a prior shift, it is also necessary to understand what the actual requirements on the model in the domain are. Sometimes, e.g., in the development of a test for a rare but dangerous disease, it might be more important to not miss actual cases than to minimize the overall risk for the real prior distribution in the population. In that case, a prior distribution might contain an equal share of positive and negative samples to boost the detection rate and the expense of more false positives. If the model is able to closely approximate the actual concept, i.e., the overall error is low, the benefit of reweighting seems to diminish. An example of this is given during the naturalistic evaluation of a prior shift in \Cref{sec:evaluation:naturalistic:breast_cancer}.

\subsection{Covariate Shift}
\label{sec:artifact:covariate_shift}

A covariate shift appears only in $X \rightarrow Y$ problems and is defined as the case where $P_s(y|x)=P_t(y|x)$ and $P_s(x) \neq P_t(x)$ \citep{moreno-torres2012}. An example of a covariate shift in \ac{ML} is learning from medical data about patients in a rural area to predict the likelihood of disease versus learning from medical data in a city hospital. In literature, this is also referred to as sample selection bias \citep{huang2006}. If the disease probability only depends on certain risk factors of the patients, this concept $P(y|x)$ will be constant in all domains. However, if we assume that the population in the rural area is older on average, this can mean that the proportion of patients with elevated risk factors is higher simply because of that. The difference compared to a prior shift is that the concept remains the same, which means that in theory the correct hypothesis can be learned from the source domain alone. However, in cases where the model used is not able to correctly represent the real concept, e.g., if we train a linear classifier to describe a non-linear problem, domain adaptation approaches can still be beneficial. This will be elaborated based on the error bound in \Cref{eq:ErrorBoundCovariate}. From the perspective of a \ac{ML} practitioner, prior shift and covariate shift can look the same for a new problem, because both result in shifts in the feature and the label space, but they require different solution procedures, and, therefore must be correctly identified.

\begin{figure}
    \centering
    \includegraphics[width=\linewidth]{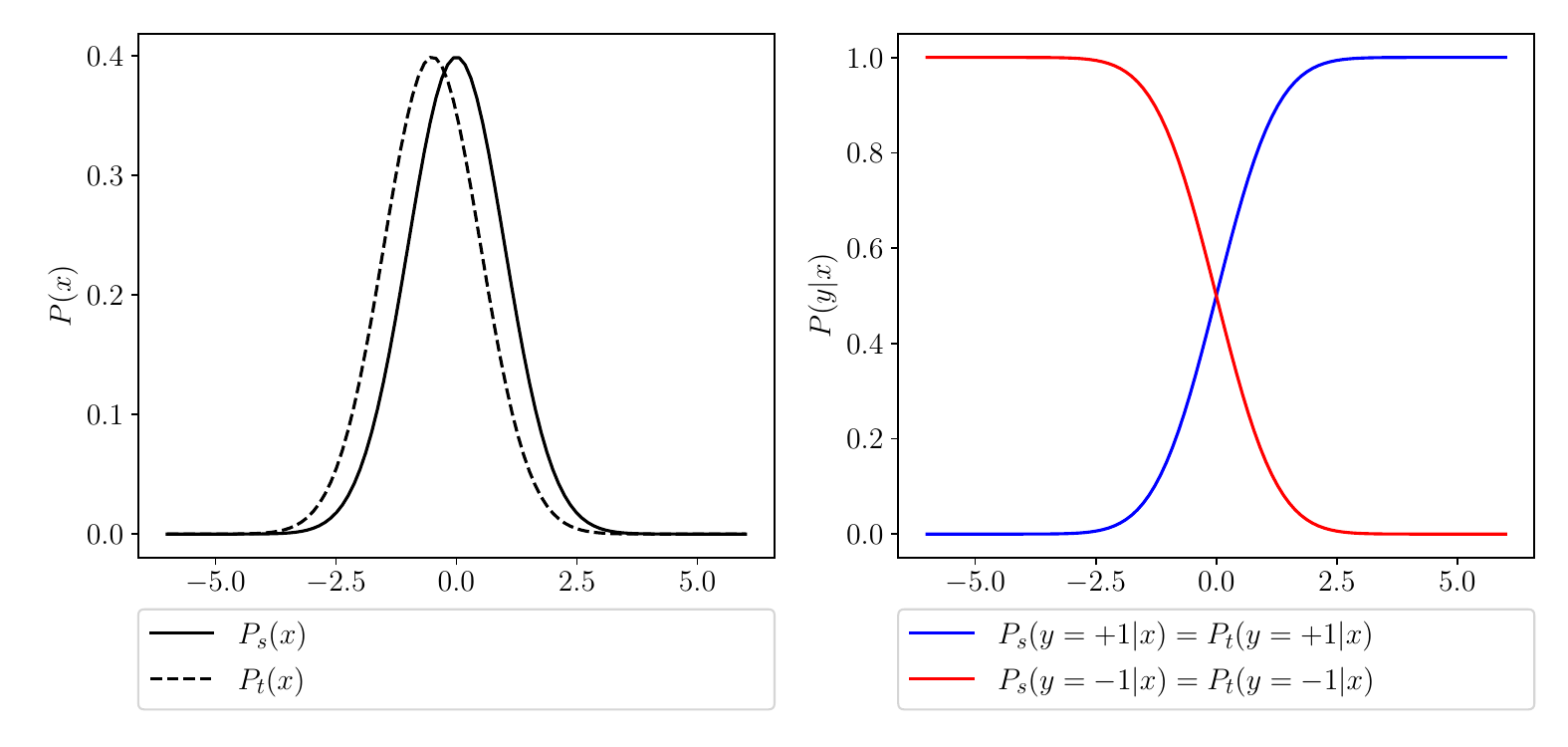}
    \caption[Covariate shift distributions.]{In $X \rightarrow Y$ systems, a \textbf{covariate shift} occurs if the distribution in the feature space (left) changes, but the concept (right) remains the same.}
    \label{fig:CovariateShift}
\end{figure}

Solution procedures of covariate shifts share the general idea with solutions to the prior shift as both scenarios can be seen as parallel cases for different causal directions. Keeping in mind that only the feature distribution changes while the concept stays constant, the target risk can be described as a decomposition similar to the prior shift scenario. 

\begin{equation}
\label{eq:CovariateWeighing}
    R_t(h)=\sum_{y\in Y} \int_{\mathcal{X}}\ell \left(h(x),y \right) \frac{\cancel{P_t(y|x)}P_t(x)}{\cancel{P_s(y|x)}P_s(x)}P_s(x,y) \,dx
\end{equation}

Even if these look almost the same, from a domain adaptation point of view there is a difference, because the weight function can be directly determined as we know the feature distribution of both source and target domain, and also the actual concept does not change between domains. What does this mean for a \ac{ML} practitioner? If the classifier of the source domain is able to correctly approximate the real concept, there is no need for domain adaptation. This is the case if the classifier has the capacity to correctly learn the real concept, e.g., a linear classifier can learn a linear concept, a quadratic classifier can learn a quadratic concept, etc. For more complex problems, it is not always clear if a classifier actually learns correctly, i.e., is well-specified. Indicators of a well-specified model are high accuracy and good generalization to new problems of the same type. In this case, the error of the source model on the target domain will asymptotically converge towards the performance on the source domain with increasing sample size.

In the case of a misspecified model, e.g., a linear classifier on a non-linear concept,  the real concept is not learned and therefore there is potential for increasing the performance by reweighting the samples. Intuitively, this allows the classifier that must make mistakes, to make the mistakes where they hurt the overall performance least. Theoretically, for a probability of $1-\delta$ for $\delta > 0$ the difference between the error of the reweighted classifier $\hat{e}_\mathcal{W}$  and the actual error on the target domain $e_\mathcal{T}$ for a classifier $h$ is defined by the following upper bound 

\begin{equation}
\label{eq:ErrorBoundCovariate}
\begin{split}
e_{\mathcal{T}}(h)-\hat{e}_{\mathcal{W}}(h)\leq \\
2^{5/4}2^{D_{2R}(P_t||P_s)/2} \sqrt[3/8]{\frac{c}{n}\log\frac{2ne}{c}+\frac{1}{n}\log\frac{4}{\delta}}
\end{split}
\end{equation}

where $D_{2R}(P_t||P_s)$ is the second-order R\'enyi divergence from \Cref{def:RenyiDivergence}, $c$ is the pseudo-dimension of the hypothesis space, and $n$ is the sample size \citep{cortes2010}. This bound shows us that for a fixed hypothesis space, more samples are needed to compensate for a higher difference between the domains. Considering the definition of the R\'enyi divergence, the importance-weighted classifier will only converge, if the expected squared weight is finite, i.e., for $w(x)=P_t(x)/P_s(x)$, it must hold that $\mathbb{E}_\mathcal{S}[w(x)^2] < \infty$ \citep{kouw2019}. For a correctly specified model, as the sample size goes to infinity, a weighted estimator will converge to the optimal solution for any fixed set of non-negative weights that sums to 1 \citep{white1981}, which includes a non-weighted estimator. 

There exist various approaches for reweighting, whereas an important differentiation is if the approach is parametrized, i.e., a certain feature distribution (e.g., a Gaussian distribution \citep{shimodaira2000}) is assumed, or non-parameterized (e.g., via kernel density estimations \citep{sugiyama2005}). Another approach is to not estimate the probability distribution in each domain but to determine the weights directly \citep{sugiyama2012}. There exists a multitude of solution approaches for correcting covariate shifts. 
A good starting point is the chapter ``Importance Weighting'' in the survey of \citet{kouw2019} about unsupervised domain adaptation. 

\subsection{Concept Shift}
\label{sec:artifact:concept_shift}

Technically, concept shift can occur in both causal directions and is therefore defined as either 
\begin{itemize}
\item $P_s(y|x)\neq P_t(y|x)$ and $P_s(x)=P_t(x)$ in $X\rightarrow Y$ problems, and
\item $P_s(x|y)\neq P_t(x|y)$ and $P_s(y)=P_t(y)$ in $Y\rightarrow X$ problems \citep{moreno-torres2012}.
\end{itemize}
Despite the similarities, these two scenarios require completely different approaches to domain adaptation. Also, it is worth noting, that in a narrow sense concept shift in $X\rightarrow Y$ problems is not a case of domain adaptation per se, because the task is changing. This violates \Cref{DomainAdaptation}. In the literature, the term concept drift is often used for this scenario \citep{widmer1996}. To avoid possible confusion, the framework slightly deviates from the definitions of \citet{moreno-torres2012} and reserves the term concept shift for $X \rightarrow Y$ scenarios, whereas a concept shift for $Y \rightarrow X$ will be called \emph{class-conditional shift} (without the loss of generality concerning regression problems) and dealt with in \Cref{sec:artifact:class_conditional_shift}.

\begin{figure}
    \centering
    \includegraphics[width=\linewidth]{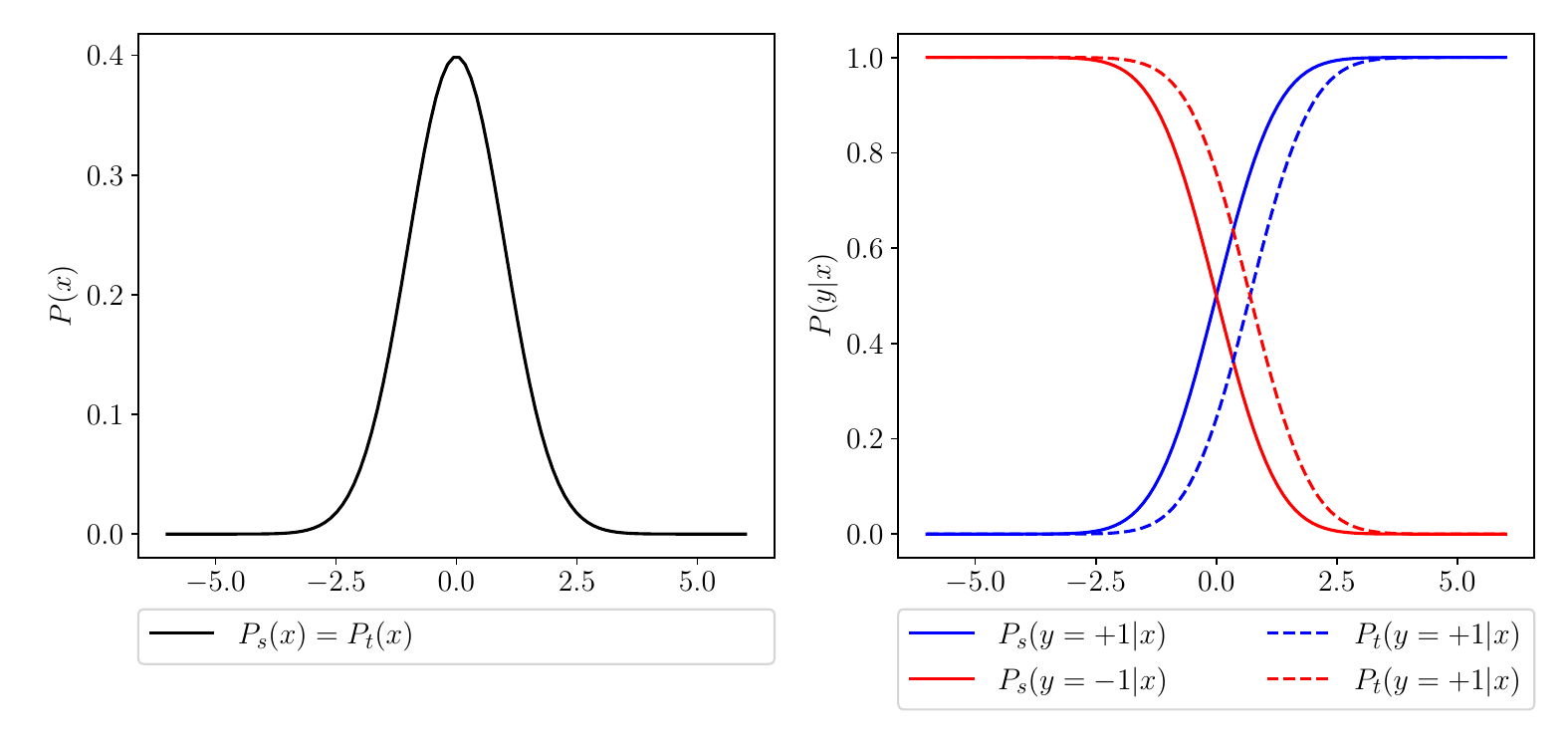}
    \caption[Concept shift distributions.]{In $X \rightarrow Y$ systems, a \textbf{concept shift} is the opposite of a covariate shift: The feature distribution (left) remains constant, but the task (right) changes in the target domain.}
    \label{fig:my_label}
\end{figure}

To illustrate concept shift, we revisit the example in the introduction of this section. In the example, the failure probability of a machine is predicted based on various sensor readings. We can assume that even if the model is trained and performs well at the moment, at some point in the future the real failure probability will change, because of wear-down in the machine. So the model that was trained at the beginning of the life cycle might underestimate the failure probability for an older machine. In a broad sense, the task \emph{predict failure probability} remains the same, however from a statistical point of view, because the parameters change explicitly, the task is also considered different. This is unlike other domain adaptation scenarios, where the task $P(y|x)$ changes implicitly, because of another change that also affects the feature space probability $P(x)$. 

\subsection{Class-Conditional Shift}
\label{sec:artifact:class_conditional_shift}

\begin{figure}
    \centering
    \includegraphics[width=\linewidth]{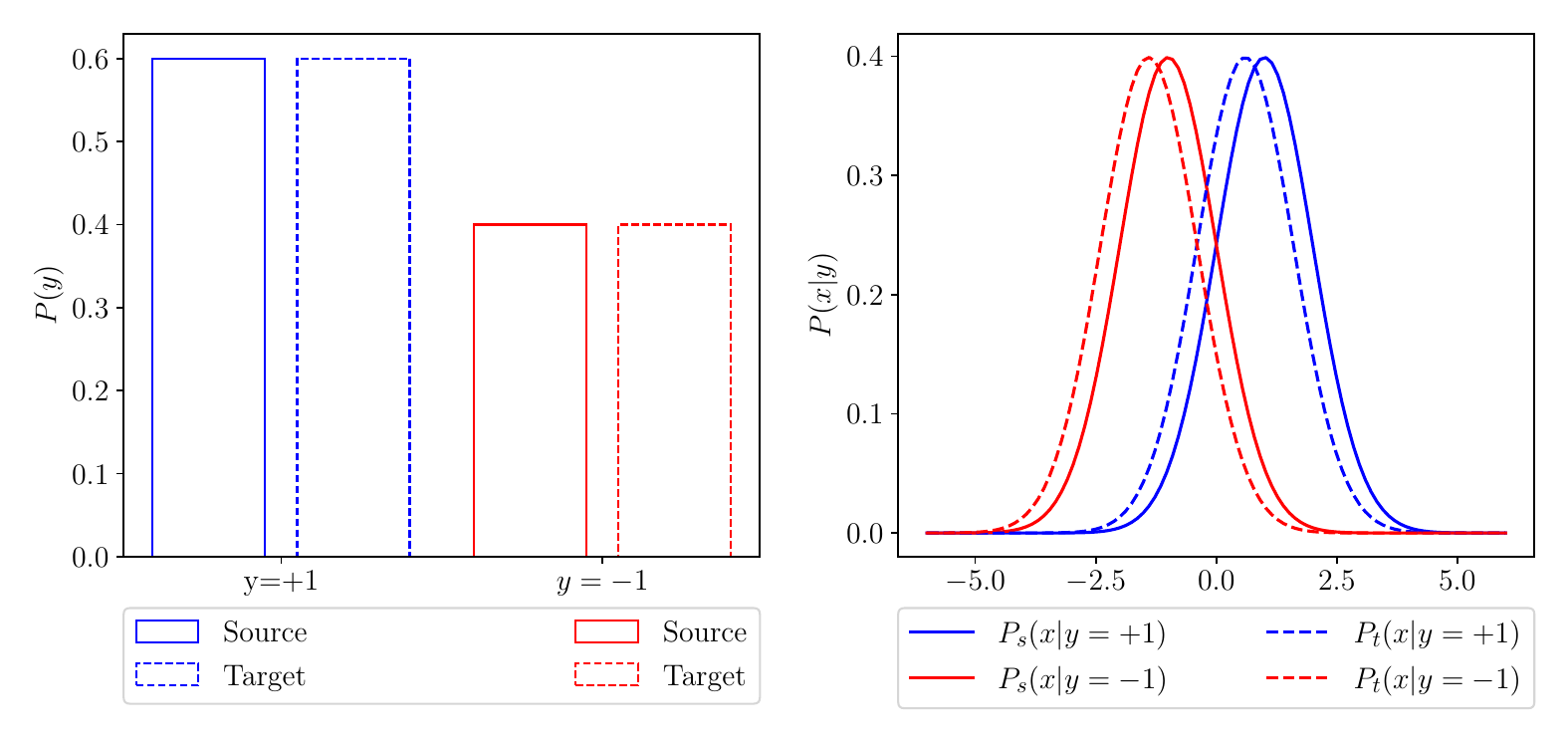}
    \caption[Class-conditional shift distributions.]{In $Y \rightarrow X$ systems, a \textbf{class-conditional shift} occurs if the change in the joint probability $P(x \cap y)$ is due to changes in the class-conditional distributions (right). The marginal probabilities of the labels (left) do not change.}
    \label{fig:ConditionalShift}
\end{figure}

In contrast to concept shift in $X \rightarrow Y$ problems, a class-conditional shift is a highly relevant scenario for domain adaptation. This type of shift is often found if observational data of the same event, but from different sources must be integrated. Remember the example of image recognition of dogs and cats from previously, but now assume that the share of dogs and cats is equal in both domains. Instead, in the target domain all pictures are taken with a different kind of camera, or maybe at night. It is immediately evident, that this will change the class-conditional $P_t(x|y)$ and it can be shown that this will also change the concept $P_t(y|x)$ and the feature distribution $P_t(x)$ as a result. Thus, we can express the relationship between the image created by a given $y$ in the source domain and the image created in the target domain via a transformation $t$ (cf. Proof in \Ref{sec:appendix:ClassConditionalShift}).

\begin{figure*}
\captionsetup[subfigure]{justification=centering}
    \centering
    \begin{subfigure}[b]{0.7\textwidth}
    \centering
    \includegraphics[width=\textwidth]{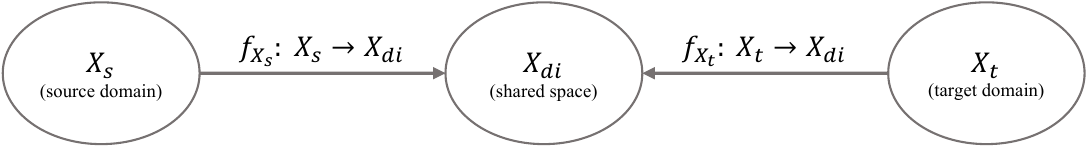}
    \caption{Shared embedding}
    \label{subfig:SharedEmbedding}
    \end{subfigure}
    \par\bigskip
    \begin{subfigure}[b]{0.7\textwidth}
    \includegraphics[width=\textwidth]{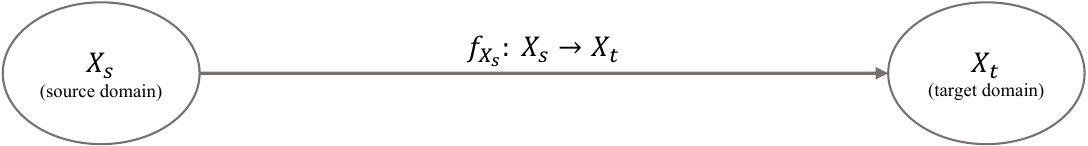}
    \caption{Domain mapping}
    \label{subfig:DomainMapping}
    \end{subfigure}
    \caption[Shared embedding versus domain mapping.]{In a shared embedding approach, both the source and the target domain are transformed into a third domain that is considered \emph{domain-invariant} (\Cref{subfig:SharedEmbedding}). In domain mapping, the source domain is \emph{directly mapped} (\Cref{subfig:DomainMapping}) to the target domain (or vice-versa) to bridge the domain difference. }
    \label{fig:SharedEmbeddingVsDomainMapping}
\end{figure*}

This means that there exists a transformation function that can be learned only in the feature space, i.e., without labels which also connects concepts of the source and the target domain. Finding this transformation function is the underlying task of all domain adaptation approaches for class-conditional shifts. \Cref{fig:SharedEmbeddingVsDomainMapping} shows that the transformation function can either map the source domain directly to the target domain or, alternatively, map both domains to a common domain-invariant space. A lower bound for the error of the latter procedure is given by \citet{gong2016}. The error requires the definition of Conditional Invariant Components: $X^{ci}$ are $d$-dimensional components of $\mathcal{X}$, obtained by transforming the data $X^{ci}=t(x)$, such that $P_s(x^{ci}|y)=P_t(x^{ci}|y)$ \citep{gong2016}. If the transformation $t(\cdot)$ is non-trivial and the target class-conditional distributions are assumed to be linearly independent of each other for any two classes, the upper bound can be expressed as \citep{kouw2019}: 
\begin{equation}
e_\mathcal{T}-e_{\hat{\mathcal{S}}}\leq J^{ci}\mathbbm{1}_{(0,\pi/2]}(\theta)+\frac{2}{\sin^2\theta}J^{ci}\mathbbm{1}_{(0,\pi/2]}(\theta)
\end{equation}
where $e_{\tilde{\mathcal{S}}}$ is the error on the transformed source data, $\mathbbm{1}$ is the indicator function, and $J^{ci}=\|\mathbb{E}_S[\phi(t(x))]-\mathbb{E}_T[\phi(t(x))]\|^2$ is the \ac{MMD} divergence between the transformed data in the source and the target domain \citep{gong2016, kouw2019}. This bound tells us that for a perfect transformation function, i.e., if $J^{ci}=0$, the difference between the two errors will also be 0. Otherwise, the difference in error depends on how well the transformation is performed for each class $c \in \mathcal{Y}$, which is expressed via the angle $\theta$ between two Kronecker delta kernels \citep{gong2016}. For too dissimilar transformed distributions, i.e., if the transformation function $t(\cdot)$ is not well selected, $\theta = \pi$ is true for the angle and the \ac{MMD} divergence cannot be used as an upper bound for the error anymore.

The theoretical validity of a solution for a class-conditional shift is given above, but we have not yet established actual solution procedures. To strike a balance between practice and theory without getting lost in the details of state-of-the-art approaches, we will briefly elaborate on established views. For deeper insights, we refer to surveys of \citet{kouw2019} for general unsupervised domain adaptation and \citet{wilson2020} for unsupervised deep domain adaptation, which describes the state of the art for domain adaptation in computer vision. To reiterate: the goal of all solution procedures in this scenario is to approximate a transformation function between the source and the target feature space. If this is done directly, it is called \emph{subspace mapping}. Usually, the details of the actual transformation are less interesting in practice because the procedure should also work for other but similar target domains. In that case, there is a common domain defined that is reachable via source and target domain with respective transformations. These procedures are commonly referred to as finding \emph{domain invariant spaces} \citep{kouw2019} or \emph{shared embeddings} \citep{murez2018}. \Cref{fig:SharedEmbeddingVsDomainMapping} illustrates the difference between the two approaches.

\begin{figure}
\captionsetup[subfigure]{justification=centering}
    \centering
    \begin{subfigure}[b]{0.49\textwidth}
    \centering
    \includegraphics[width=\textwidth]{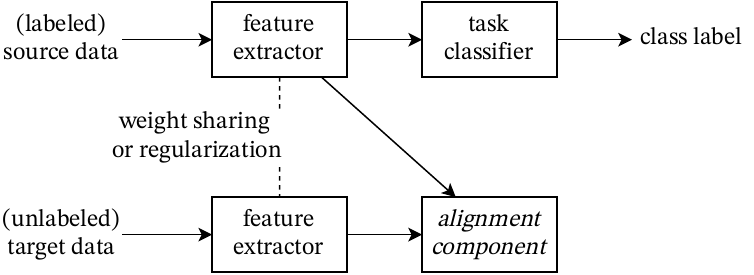}
    \caption{Training}
    \label{subfig:FeatureExtractorTraining}
    \end{subfigure}
    \bigskip
    \medskip
    \hfill
    \begin{subfigure}[b]{0.49\textwidth}
    \includegraphics[width=\textwidth]{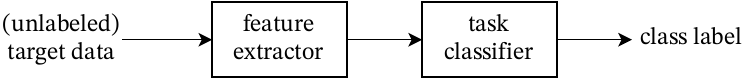}
    \caption{Testing}
    \label{subfig:FeatureExtractorTesting}
    \end{subfigure}
    \caption[Transformation function as a feature extractor.]{A possible setup for domain adaptation is to simultaneously train the feature extractor and the classifier based on transformed source data. During training, it is important to regularize the feature extractor so that it not only minimizes the divergence between the domains but also extracts those features that are important for the task at hand. The figure is borrowed from the domain adaptation survey of \citet{wilson2020}.}
    \label{fig:FeatureExtractor}
\end{figure}

\citet{wang2018} and \citet{wilson2020} surveyed deep domain adaptation approaches and identified possible classifications for approaches.  Many approaches use \acp{GAN} in order to connect the task of finding a meaningful domain invariant space with the actual task at hand, e.g., a classification \citep{wilson2020}. For a neural network, the transformation function $t(\cdot)$ can be interpreted as a feature extractor, that is set up in front of the actual classifier as shown in \Cref{fig:FeatureExtractor}.

While \acp{GAN} are a popular choice today, other recent approaches also yielded promising results. An example of this is given by \citet{french2018}. Here, temporal self-ensembling is used to align the data set distributions, i.e., to implicitly learn the transformation function $t(\cdot)$. The bad news for \ac{ML} practitioners is that for now, there is no known way to precisely define a priori which approach is the best for a given problem, as this would require perfect knowledge about the prior distribution in both domains \citep{kouw2019, wolpert1996}. An advisable approach is to rely on surveys that summarize well-performing state-of-the-art approaches and select promising candidates at the discretion of expert knowledge about the underlying mechanics and the respective domain characteristics.

\subsection{General Data Set Shift}
\label{sec:artifact:general_shift}
All combinations of shifts that are not captured in the first four rows of \Cref{tab:scenarios} are defined as general data set shifts or simply other types of data set shifts, this includes mainly \citep{moreno-torres2012}: 
\begin{itemize}
\item $P_s(y|x)\neq P_t(y|x)$ and $P_s(x)\neq P_t(x)$ in $X \rightarrow Y$ problems, and
\item $P_s(x|y)\neq P_t(x|y)$ and $P_s(y)\neq P_t(y)$ in $Y \rightarrow X$ problems. 
\end{itemize}

Another way to look at general data set shifts is to see these scenarios as combinations of the two basic scenarios depending on the causality of the system. \citet{moreno-torres2012} states that possible reasons for the rarity of these types of shifts in current literature are that they appear not as often as the basic variants and that they simply are too hard to solve. At least the first point does not hold for domain adaptation, because general data set shifts can easily occur in practice. For example, let us revisit the problem of discriminating between pictures of cats and dogs with a binary classifier for the last time. One can easily imagine a scenario in which the target differs from the source in both the prior probabilities, i.e., the different share of cats versus dog images, and in the class conditionals, i.e., different cameras or picture qualities. The second point, namely the difficulty of these problems, holds still true.

In the case of a general data set shift, none of the solution procedures above can be expected to work as before, because for each of them at least one assumption concerning the equality of certain distributions, is violated. For purely statistical methods, e.g., reweighting in the case of a covariate or prior shift, this is completely true. For a class-conditional shift, it is possible that depending on the extent of the prior shift, certain approaches will still deliver adequate results. As a suggestion, for general data set shifts in $Y\rightarrow X$ scenarios, the same solution procedures as for a class-conditional shift can be applied, but following the line of argumentation from the error bound in \Cref{eq:LowerBoundErrorGeneralShift}, a performance drop can occur, because the real concept cannot be learned from aligning the feature space in source and target domain. Intuitively, a reasonable assumption is that the more accurate the transformation function and the smaller the prior shift, the better the performance of the approach.

\citet{zhao2019} formalized this reasoning and constructed an upper and lower bound for learning domain invariant features, i.e., domain adaptation for class-conditional shifts, under different label distributions. Consider the following upper bound. 

\begin{equation}
\label{eq:UpperBoundErrorGeneralShift}
\begin{aligned}
e_{\mathcal{T}}(h) \leq e_{\hat{\mathcal{S}}}(h) &+d_{\tilde{\mathcal{H}}}(\hat{D}_\mathcal{S}, \hat{D}_\mathcal{T})+ 2 \text{Rad}_\mathcal{S}(\mathcal{H})+4\text{Rad}(\mathcal{H}) \\
&+ \min\{\mathbb{E}_{D_\mathcal{S}}[|f_\mathcal{S}-f_\mathcal{T}|], \mathbb{E}_{D_{\mathcal{T}}}[|f_\mathcal{S}-f_\mathcal{T}|]\} \\
&+ O(\sqrt{\log{(1/\delta)/n})}
\end{aligned}
\end{equation}

where $\tilde{\mathcal{H}} := \{\text{sgn}(|h(x)-h'(x)|-t|h,h' \in \mathcal{H}, t\in [0,1]\}$ and 
$\text{Rad}_\mathcal{S}(\mathcal{H})$ denotes the empirical Rademacher complexity. The bound is not discussed in detail here. More important is that three major components of the upper bound can be identified: the part of the error that belongs to the domain adaptation setting, e.g., the empirical source error and the shift between the label distributions, the part of the error that stems from the complexity measures in the hypothesis spaces and the error that is caused by only having finite samples \citep{zhao2019}.  

The lower bound for the difference in the errors is also interesting because it describes which performance can be expected in the best case for any domain adaptation based on domain invariant representations: 

\begin{equation}
\label{eq:LowerBoundErrorGeneralShift}
e_\mathcal{S}(h \circ g) + e_\mathcal{T}(h \circ g) \geq \frac{1}{2}(d_{JS}(D^Y_\mathcal{S}, D^Y_\mathcal{T})-d_{JS}(D^Z_\mathcal{S}, D^Z_\mathcal{T}))^2
\end{equation}

In this equation, following the notation of \citet{zhao2019}, $d_{JS}$ is the Jensen-Shannon distance and the functions $g$ and $h$ represent the domain-invariant learning as a Markov chain:

\begin{equation}
\label{eq:MarkovChain}
X \overset{g}{\rightarrow}Z \overset{h}{\rightarrow} \hat{Y}
\end{equation}

where $\hat{Y}=h(g(X))$ is the predicted random variable of interest, $Z=g(X)$ is the domain invariant feature space, and $X$ is the original domain. This lower bound shows that for sufficiently different label distributions, minimizing the source error $e_\mathcal{S}(h\circ g)$ and aligning the distributions in the domain invariant feature space by minimizing $d_{JS}(D^Z_\mathcal{S}, D^Z_\mathcal{T})$ will only increase the target error \citep{zhao2019}. The error bounds show that for a general data set shift, there is no guarantee for a domain adaptation approach to work if there is a sufficiently large difference in the marginal label distributions, or if the domains are too different, if the real concept is too complex or if the finite samples are too limited. 

Many articles that present new algorithms do not state the performance in the case of unbalanced classes so it is difficult to make exact statements. A possible direction of research is to evaluate the performance of the state-of-the-art approaches for problems that are categorized by the here presented framework. 
Lastly, it should be noted that not all solution procedures necessarily rely on risk minimization and some approaches are naturally robust against certain types of shifts. Support vector machines \citep{cortes1995} are a popular example because they maximize the margin between a separating hyperplane, which makes them more robust against outliers and---depending on the kernel---more robust against some forms of domain shifts. In the technical evaluation for general data set shifts (\Cref{sec:evaluation:artificial}), we present an example of this.
For domain adaptation concerning general data set shift problems, an advisable course of action remains to leverage domain knowledge and exploit connections between the source and target domain that are not necessarily captured by the statistical properties available in the problem itself. In a sense, it might be helpful to approach domain adaptation as its own \ac{ML} problem rather than an add-on to the original task. 

\subsection{Scenario Determination}
\label{sec:artifact:scenario_determination}
From a practitioner's point of view, one of the first questions is likely to be whether or not a given problem is suitable for domain adaptation at all. While there is never a guarantee that a specific domain adaptation approach yields the desired results, at the start of a project it can be helpful to have an indicator if the problem fits at least in the correct class of problems. This is why this section elaborates on how to determine if a \ac{ML} problem belongs to one of the scenarios described in \Cref{tab:scenarios}. For the general case, domain adaptation is probably too complex to ever find a closed set of statistical tests that can precisely pinpoint which approach is suitable. Although domain knowledge remains the most important factor in determining which scenario a problem belongs to, some of these scenarios can also be determined by a statistical test. This, however, usually assumes that the causality of the system is known. For many problems, the determination of the causality can be a trivial task, but in some instances, it is possible that the connection between $X$ and $Y$ is not clear. In that case, \emph{causal research} \citep{pearl2009} must be performed beforehand.

If a $Y \rightarrow X$ causality is determined, there exists a statistical test for identifying if an expectation-maximization-based class reweighting as proposed by \citet{saerens2002} has a positive influence in the case of a prior shift. Identifying a class-conditional shift is more difficult because of the missing knowledge about the labels in the target domain. It is advisable to mainly rely on domain knowledge to identify these scenarios, but a relaxation of the unsupervised assumption makes it possible to include common hypothesis testing from statistics as an indicator. By manually labeling some samples from the target domain, one can test if the samples are drawn from the same distribution, e.g., via the two-sample Kolmogorov-Smirnov test \citep{masseyjr1951}. If the test suggests it is likely that the label distributions are identical, but the performance of the source classifier is not identical on the target domain, then it is possible that a class-conditional shift has occurred.

In $X \rightarrow Y$ problems, a strong indicator of a concept shift is that $P(x)$ remains the same in both domains. On the other hand, $P(x)$ changes during a covariate shift or a general data set shift. For lower dimensional problems the feature distribution can be parametrically or non-parametrically estimated and the divergence quantified via a suitable divergence measure, e.g., Kullback-Leibler divergence. If no change in the feature distribution is observed, but an otherwise well-performing model shows a drop in accuracy, it is likely that a concept shift has occurred. If on the other hand, there is a difference in the feature distribution, the further procedure depends on the fitness of the model. If the model is correctly specified and of suitable complexity to approximate the real concept, then the source classifier should perform comparably well on the target domain and domain adaptation is not necessary. If however, the model is misspecified, we can observe a drop in performance on the target domain. Domain adaptation approaches can positively influence performance in this case. Independently of that, domain knowledge can help immensely in identifying the correct scenario. \Cref{tab:ScenarioDetermination} summarizes the presented determination procedures.

\begin{table*}
\caption[Scenario determination]{Procedures for scenario determination}
    \label{tab:ScenarioDetermination}
    \centering
    \begin{tabularx}{\linewidth}{l l X}
    \toprule
    \multicolumn{2}{l}{Scenario} & Determination procedure  \\ \cmidrule{1-2}
         Causality& Shift& \\
         \midrule
         $ Y \to X$ & Prior & Determine shift in priors via domain knowledge, semi-supervised hypothesis testing (e.g., Kolmogorov-Smirnoff), maximum-likelihood hypothesis testing, e.g., \citet{saerens2002}, or application of solution procedure with a beneficial outcome.
\\
& Class-cond. & Determine shift in class conditionals via domain knowledge, exclusion of prior shift, or application of solution procedure with a beneficial outcome. \\
\midrule
$X \to Y$ & Covariate & Determine shift in feature space via observation, e.g., by hypothesis testing or distribution divergence, or by application of solution procedure with a beneficial outcome. \\
& Concept & Determine a concept shift by excluding a covariate shift, but still observing a performance drop for an otherwise well-generalizing model in the target domain. \\
\midrule
Unspecified & General & Determine a general shift via domain knowledge and combination of the procedures above, if applicable.\\
    \bottomrule
    \end{tabularx}
\end{table*}

\section{Evaluation}
\label{sec:evaluation}
\begin{figure}
    \centering
    \includegraphics[width=0.7\linewidth]{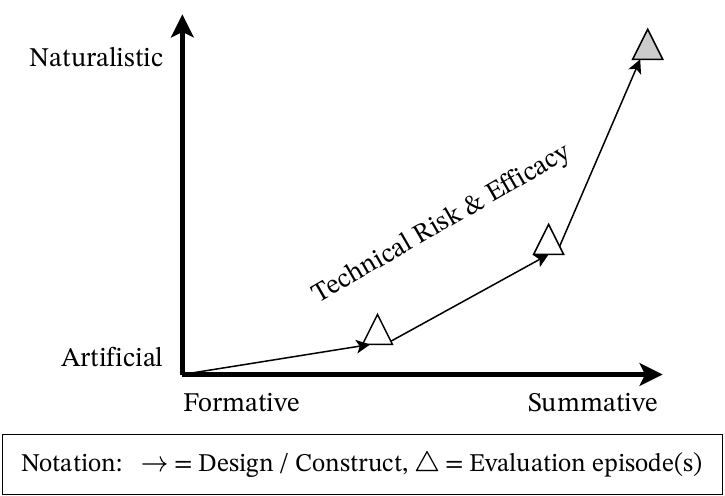}
    \caption[Evaluation episodes.]{In the technical risk \& efficacy strategy proposed by \citet{venable2016}, early artificial testing guides the development process. For the overall evaluation of the framework, three episodes are conducted. The third \ac{EE} (gray triangle) under real-life conditions was performed with a limited-scope prototype implementation.}
    \label{fig:TechnicalRiskAndEfficacy}
\end{figure}

Following the framework for evaluating \ac{DSR} by \citet{venable2016}, the strategy of evaluation is based on the goals and risks associated with the artifact. For the presented framework, the major risks are that it is not perceived as useful by practitioners or that domain adaptation proves too complex in itself so it is not possible to simplify the topic sufficiently for an expert system to work. At the same time, testing actual domain adaptation approaches is costly because the implementation of state-of-the-art algorithms for novel data sets is often associated with significant manual effort. As a result, the evaluation of the artifact uses a two-episode \emph{technical risk \& efficacy} approach as illustrated in \Cref{fig:TechnicalRiskAndEfficacy}.

In the first descriptive \ac{EE}, the application of \citet{moreno-torres2012} data shift approaches for domain adaptation are tested to create a solid theoretical foundation and formatively support the further development of the framework. In the second observational \ac{EE}, known problems from the literature are classified, and the presented solutions are analyzed using the framework. This is done to evaluate the performance of the artifact in a more naturalistic scenario and subsequently permit statements about the usefulness and explanatory power of the framework. In a third \ac{EE}, we conduct an experimental study with 50 domain experts in a between-subject design to test the practical applicability, usefulness, and intention to use.

\subsection{Evaluation Episode 1}
\label{sec:evaluation:artificial}
\begin{sidewaysfigure*}
    \centering
    \includegraphics[width=\linewidth]{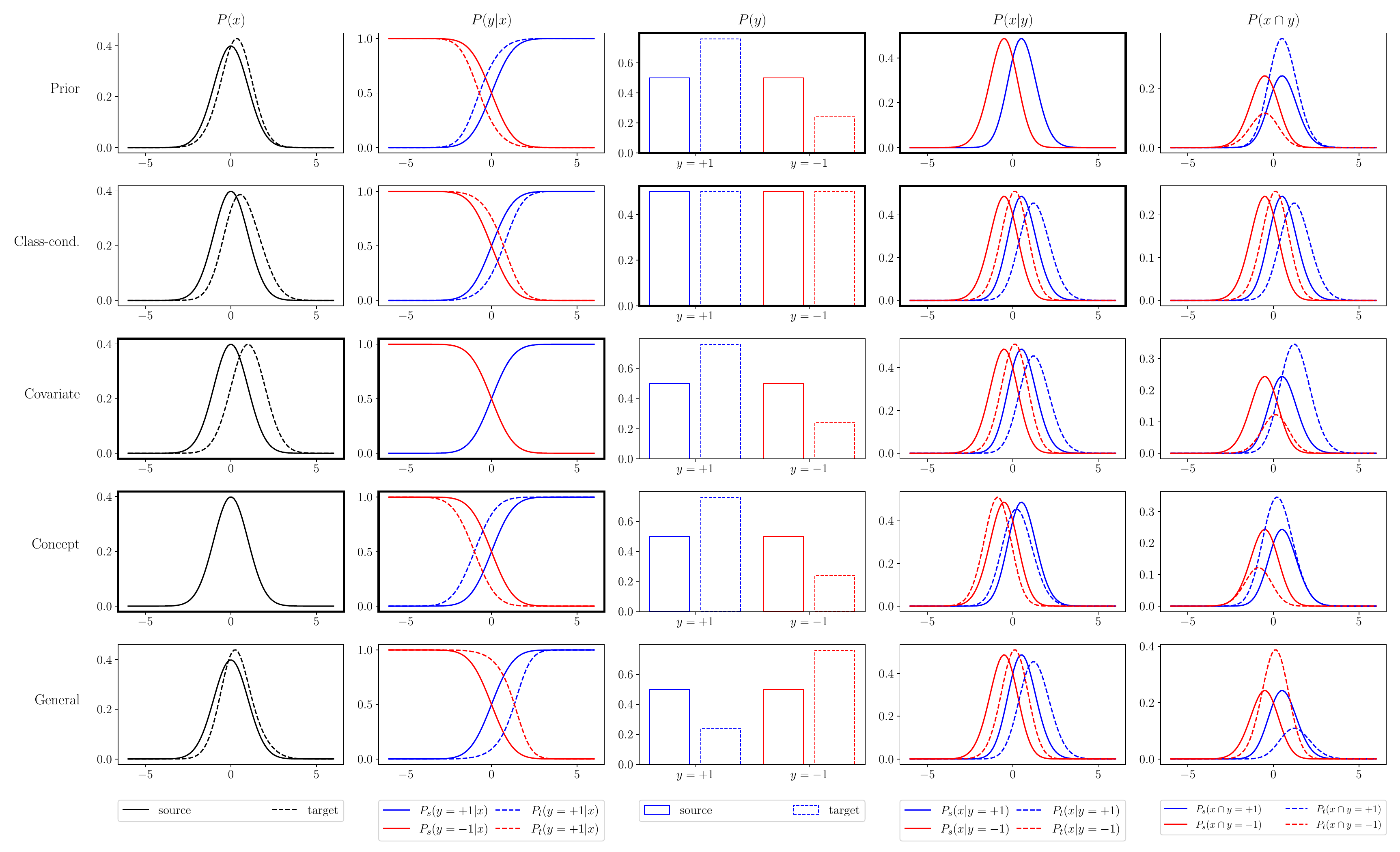}
    \caption[Overview of possible domain adaptation scenarios.]{The overview of possible domain adaptation scenarios shows the occurring shifts for a simple one-dimensional example. Attributes that are given by the respective definition are marked by thick borders.}
    \label{fig:ScenarioOverview}
\end{sidewaysfigure*}

\sloppy In the first \ac{EE}, the four actual domain adaptations scenarios: prior shift, covariate shift, class-conditional shift, and general data set shift are mathematically evaluated to confirm the inferred distribution divergences in \Cref{tab:scenarios} are correct. Afterwards, we create exemplary data sets for each scenario, if necessary, confirm their properties with non-parameterized kernel density estimations and test whether or not the predictions of the framework hold. \Cref{fig:ScenarioOverview} illustrates the occurring shifts for all scenarios, whereas shifts (or non-shifts) that are required by the definition are marked by a thick border around the respective plot. For simplicity, the example used is a one-dimensional binary classification problem, in which the ground truth feature distribution is standard normal, and the concept is given by the cumulative distribution function for $y=+1$ and the inverse cumulative distribution for $y=-1$. A qualitative analysis of the scenarios is enabled by introducing shifts as defined by \citet{moreno-torres2012}. \Cref{fig:ScenarioOverview} is meant to demonstrate the statistical relationships in the different scenarios and is no realistic example. Therefore the causality of the system is not further defined, and the exact shifts are omitted. 

\begin{figure}
    \centering
    \includegraphics[width=\linewidth]{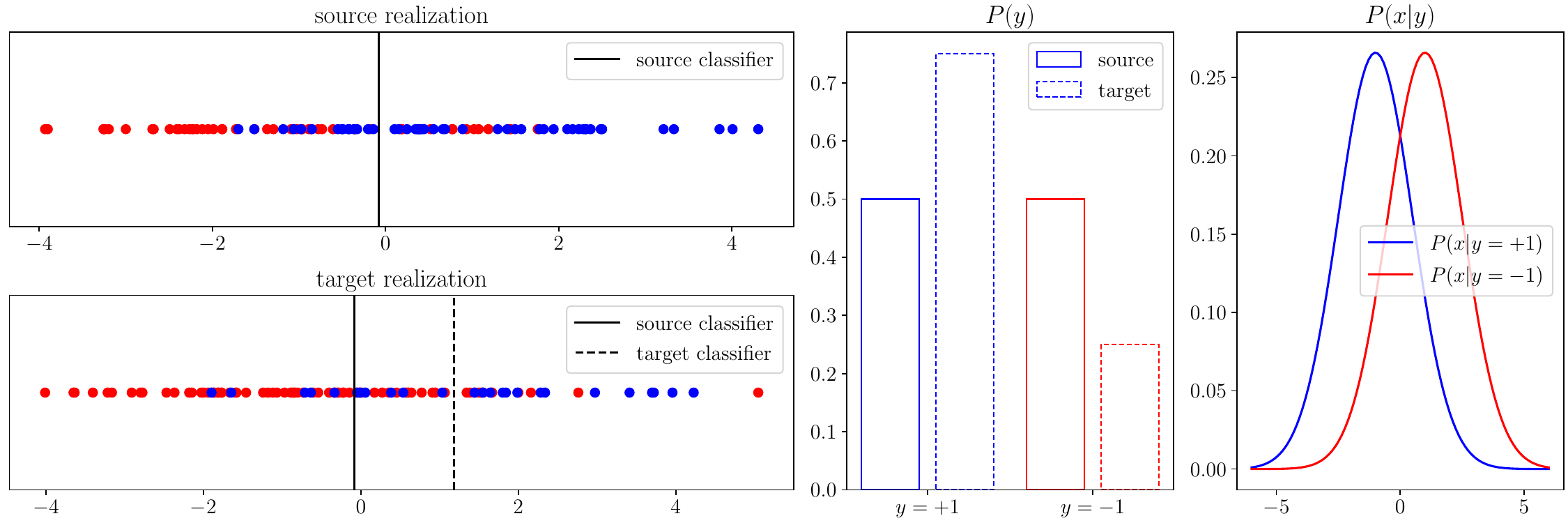}
    \caption[Prior shift evaluation.]{For a $Y \rightarrow X$ scenario, a prior shift in the target domain results in a different concept being learned by a linear classifier. In this example the prior probabilities are $P_s(y=+1) = P_s(y=-1) = 0.5$ for the source domain and $P_t(y=+1)=0.75, P_t(y=-1)=0.25$ in the target domain. The class-conditional probabilities are $P_s(x|y=+1)=P_t(x|y=+1) \sim \mathcal{N}(\mu=-1, \sigma=1.5)$ and $P_s(x|y=-1)=P_t(x|y-1) \sim \mathcal{N}(\mu=+1, \sigma=1.5)$ for both domains.}
    \label{fig:PriorEval}
\end{figure}

In a \emph{prior shift} scenario, the class conditionals remain the same, $P_s(x|y)=P_t(x|y)$, while the prior distribution differs between the source and target domain $P_s(y)\neq P_t(y)$. Following the sentence of total probability, we can show that the posterior distributions, i.e., the feature space, must also differ. The difference is propagated towards the concept and the joint probability, analogous to \Cref{eq:TotalProbability}.  \Cref{fig:PriorEval} shows the realization for a prior shift in a one-dimensional classification task. Although this example is most certainly too simple to resemble a realistic problem, it helps to understand the implications of a prior shift for learning the right concept. 
For a non-adaptive classifier, the framework predicts that for sufficiently huge differences in the prior distributions, the performance of the classifier will degrade as the real concept is different in the target domain. \Cref{tab:PriorEval} shows the performance of a linear classifier that is trained in the source domain versus the performance of a classifier that could be trained if the labels in the target domain were known. It can be seen that the classifier trained on the source can not achieve the same performance as the classifier trained on the target data with target labels. This motivates the use of domain adaptation for prior shifts. 

\begin{table}
\caption[Prior evaluation accuracy.]{Although the source classifier achieves similar performance on both the target and source domain in this example, a comparison shows that higher performance is possible in the target domain if the right concept is learned. The values are averaged for $n=10$ simulation runs and the sample size in each simulation run was 100. Otherwise, the setup is identical to the one in \Cref{fig:PriorEval}.}
\label{tab:PriorEval}
\centering
\begin{tabular}{@{}llll@{}}
\toprule
Prior shift       & & \multicolumn{2}{l}{Accuracy} \\ \midrule
Domain       &  $P(y=+1)$        & SVM (source)                & SVM (target)                 \\
\midrule
Source & 0.5       & 0.75                   & 0.64                   \\
Target & 0.75      & 0.74                   & 0.80                   \\ \bottomrule
\end{tabular}
\end{table}

\begin{figure}
    \centering
    \includegraphics[width=\linewidth]{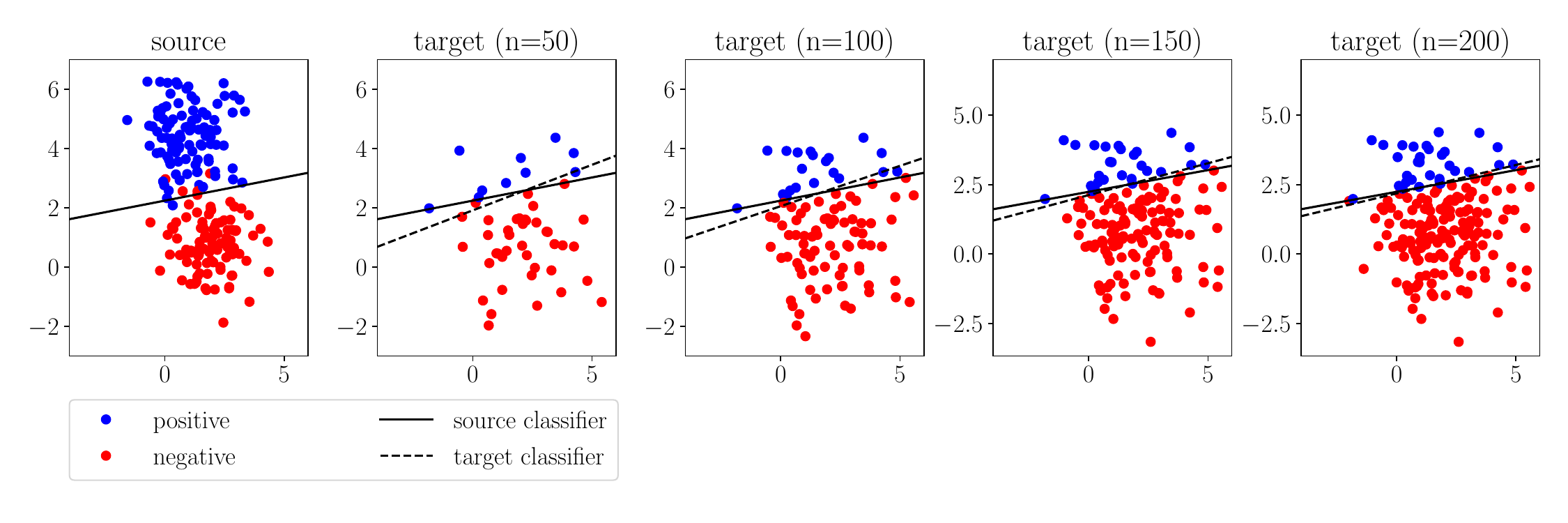}
    \caption[Covariate shift and correctly specified model.]{For a covariate shift, the correct concept can still be learned if enough sample data is available and if the model is correctly specified, i.e., is able to correctly approximate the real nature of the concept. }
    \label{fig:CovariateEval}
\end{figure}

A \emph{covariate shift} shows equal concepts $P_s(y|x)=P_t(y|x)$, but different posterior distributions $P_s(x) \neq P_t(x)$. Following the same reasons as for prior shift, we can show that the joint, class-conditional and prior probabilities each change depending on the covariate shift. For a correctly specified classifier, i.e., a classifier that has the capacity to fully capture the real concept, the framework predicts that the real concept can be learned if enough samples are provided. \Cref{fig:CovariateEval} shows this for a linear concept and a covariate shift along the y-axis for two-dimensional data. Domain adaptation can still be helpful, if the model is misspecified, i.e., if it does not have the capacity to correctly learn the real concept. An example of this is given in the naturalistic \ac{EE} for the prediction of the heart disease probability based on age and cholesterol (\Cref{sec:evaluation:naturalistic:heart_disease}). In this case, the performance of a classifier does (a) not necessarily improve with a higher number of provided samples, but (b) improves if a reweighting approach modifies the importance of the samples.

In a \emph{class-conditional shift} scenario, we observe a change in the class-conditional probabilities $P_s(x|y)\neq P_t(x|y)$ between source and target domain, while the prior distributions remain equal $P_s(y)=P_t(y)$. The other probabilities can be inferred using Bayes' theorem and the law of total probability analogously to the prior or covariate shift case. The transformation function $t(\cdot)$ is expected to translate to the feature space (cf. \Ref{sec:appendix:ClassConditionalShift}). Using kernel density estimation and a divergence measure, it can be shown that a divergence in the class conditionals creates a proportional divergence in the feature space. To test this assumption, the example of \Cref{fig:PriorEval} is modified, so that the priors now are equal in both domains and the change in the class-conditional probabilities is expressed via a shift function $t(x) = x + b$ so that $P_s(x|y)=P_t(x+b|y)$. \Cref{tab:KullbackLeiblerClassConditional} shows the Kullback-Leibler divergence in $P(x)$ and $P(x|y)$ for $b \in \{0,0.2,0.4,...,1.8,2.0\}$.  
It is left to state that if the similarity of domains is incidental and no meaningful connection between the class conditionals exists, e.g., if the shift is due to random noise, a transformation function can still be learned given a set of samples, but will not be reliable for other samples. 

\begin{table*}
    \caption[Kullback-Leibler divergence for a class-conditional shift.]{The Kullback-Leibler divergence for a class-conditional shift in a binary decision problem with one feature dimension shows the proportionality of shifts in the class conditionals $P(x|y)$ and in the feature space $P(x)$. The source class conditional is given by 
    $P_s(x|y=+1) \sim \mathcal{N}(\mu=1), P_s(x|y=-1) \sim \mathcal{N}(\mu=-1)$.}
    \label{tab:KullbackLeiblerClassConditional}
    \centering
    \small
    \begin{tabular}{l l l}
    \toprule
        Transformation & \multicolumn{2}{l}{Kullback-Leibler divergence} \\ \cmidrule{2-3}
         $P_t(x+b|y)=P_s(x|y)$ & $D_{KL}\big(P_s(x|y)||P_t(x|y)\big)$ & $D_{KL}\big(P_s(x) || P_t (x)\big)$ \\ 
         \midrule
         $b=0.0$& 0.00 & 0.00 \\
         $b=0.2$& 0.02 & 0.01 \\
         $b=0.4$& 0.08 & 0.04 \\
         $b=0.6$& 0.18 & 0.10 \\
         $\vdots$ & $\vdots$& $\vdots$\\
         $b=1.6$ & 1.28 & 0.73 \\
         $b=1.8$ & 1.62 & 0.94 \\
         $b=2.0$ & 2.00 & 1.17 \\
    \bottomrule
    \end{tabular}

\end{table*}

\begin{figure}
    \centering
    \includegraphics[width=0.9\linewidth]{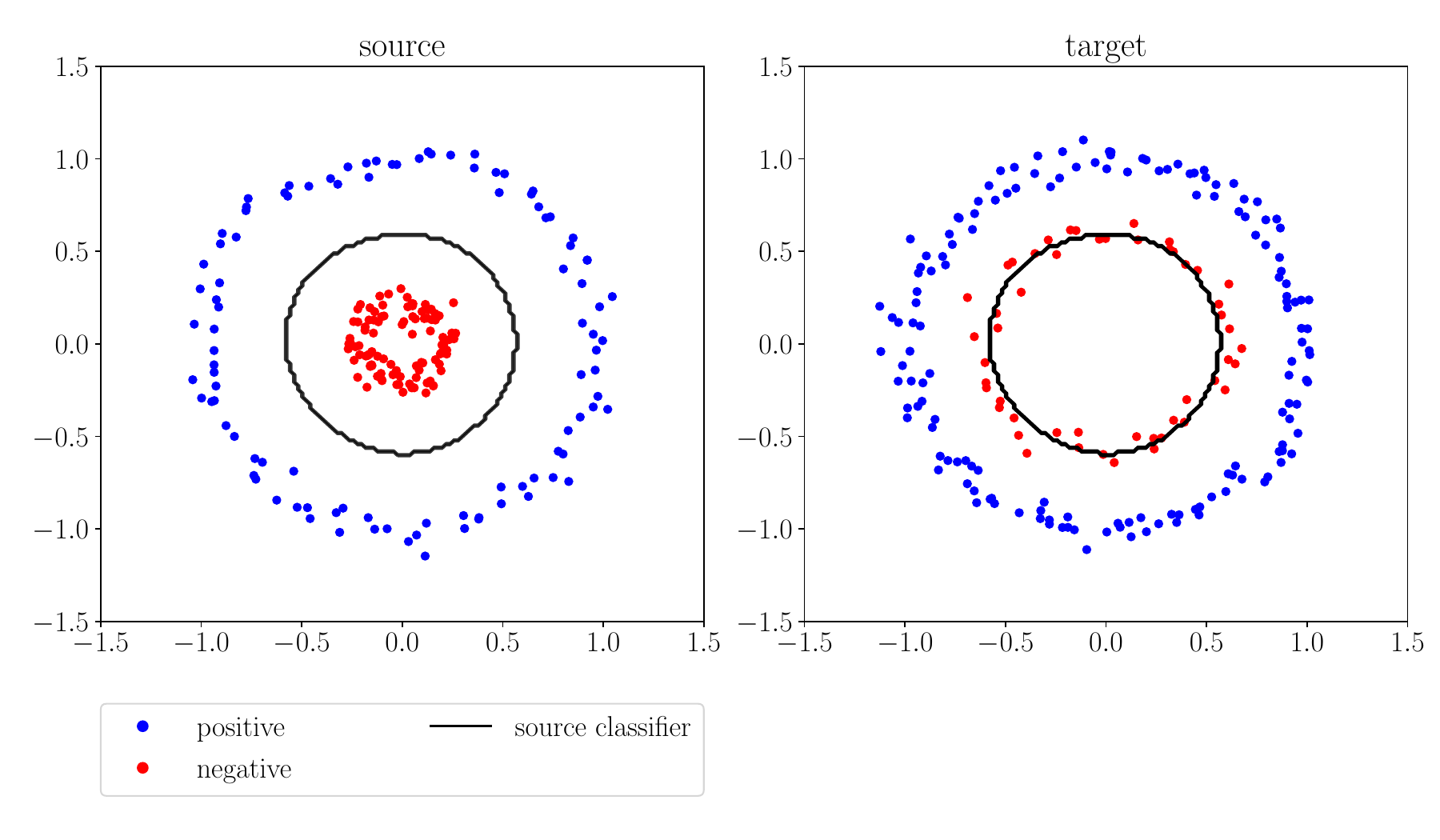}
    \caption[General shift benign case.]{Benign case of a $Y \to X$ general shift. Although the source classifier is not optimal for the target domain, the resulting error is relatively small because (a) the classifier is robust for the class-conditional shift, and (b) fewer negative instances occur, which are responsible for a large portion of the overall error.}
    \label{fig:GeneralShiftEvaluation}
\end{figure}

In the case of a \emph{general data set shift}, we only evaluate the $Y \rightarrow X$ scenario, based on the exclusion of concept drift as a domain adaptation problem explained in \Cref{sec:artifact:concept_shift}. A core statement for general data set shifts is that the performance of a model depends heavily on its implementation details. \Cref{fig:GeneralShiftEvaluation} shows a case of general data set shift for a two-dimensional feature space. The source classifier is a non-adaptive support vector machine using a \ac{RBF} kernel. In the target distribution both a prior shift, i.e., fewer negative samples than in the source, and a class-conditional shift, i.e., the negative samples are shifted away from the circle center, are introduced. This would normally result in a significant performance drop. But the source classifier performs comparatively well because of two reasons: (a) the decision boundary is still valid for all positive instances and some negative ones as the SVM maximized the margin between two classes, and (b), the proportion of negative instances has declined relative to the source so that the error in this class contributes less to the overall error. Depending on the complexity of the domain adaptation and the complexity of the trained model, these interactions can become challenging to comprehend, which means that in many cases it is hard to estimate a priori which domain adaptation approach will yield beneficial results.  

\subsection{Evaluation Episode 2}
\label{sec:evaluation:naturalistic}
A more naturalistic evaluation of the framework is achieved by using real problems from literature and applying the domain adaptation framework. We select well-established and intuitive problems, which are presented in literature alongside empirical results for at least one domain adaptation approach: linear classification of heart disease probability based on health data, detection of breast cancer in the description of cell images via a neural network, and the popular computer vision domain adaptation task consisting of images of numbers from the MNIST, USPS, and SVHN data sets. The presented examples focus on classification tasks for result interpretability. Notably, our approach seamlessly extends to regression tasks due to the absence of specific constraints on the learned concept.

\subsubsection{Heart Disease}
\label{sec:evaluation:naturalistic:heart_disease}
In this data set, created by medical researchers including \citet{steinbrunn1988}, and picked up by \citet{kouw2019} in their review of domain adaptation methods, a classifier should be trained to predict whether or not a patient will develop heart disease. Originally the refined data set contains 14 attributes from 4 clinics worldwide. In their paper \citet{kouw2019} only use the two attributes cholesterol level and age from the hospitals at Long Beach in California, USA, and Budapest, Hungary. 

First, the framework requires determining the scenario based on the causality and other attributes of the problem: Because the task is to predict if a patient \emph{will} develop a heart disease, the heart disease ($Y$) cannot be the cause of the observed attributes ($X$). Therefore we are in a $X \rightarrow Y$ scenario. From there on, it can only be one of three domain adaptation problems: covariate shift, concept shift, or a combination of those. Statistically, we can see a divergence in the feature distribution between the two data sets and therefore eliminate a pure concept shift. Without target labels, we must rely on domain knowledge to decide if a concept change is plausible. Medically, there are no structural differences between humans in Budapest and in California, so it is reasonable to assume that the real concept remains the same. The working assumption, therefore, is that we deal with a \emph{covariate shift}. 

\begin{figure*}
\captionsetup[subfigure]{justification=centering}
    \centering
    \begin{subfigure}[b]{0.75\textwidth}
    \centering
    \includegraphics[width=\textwidth]{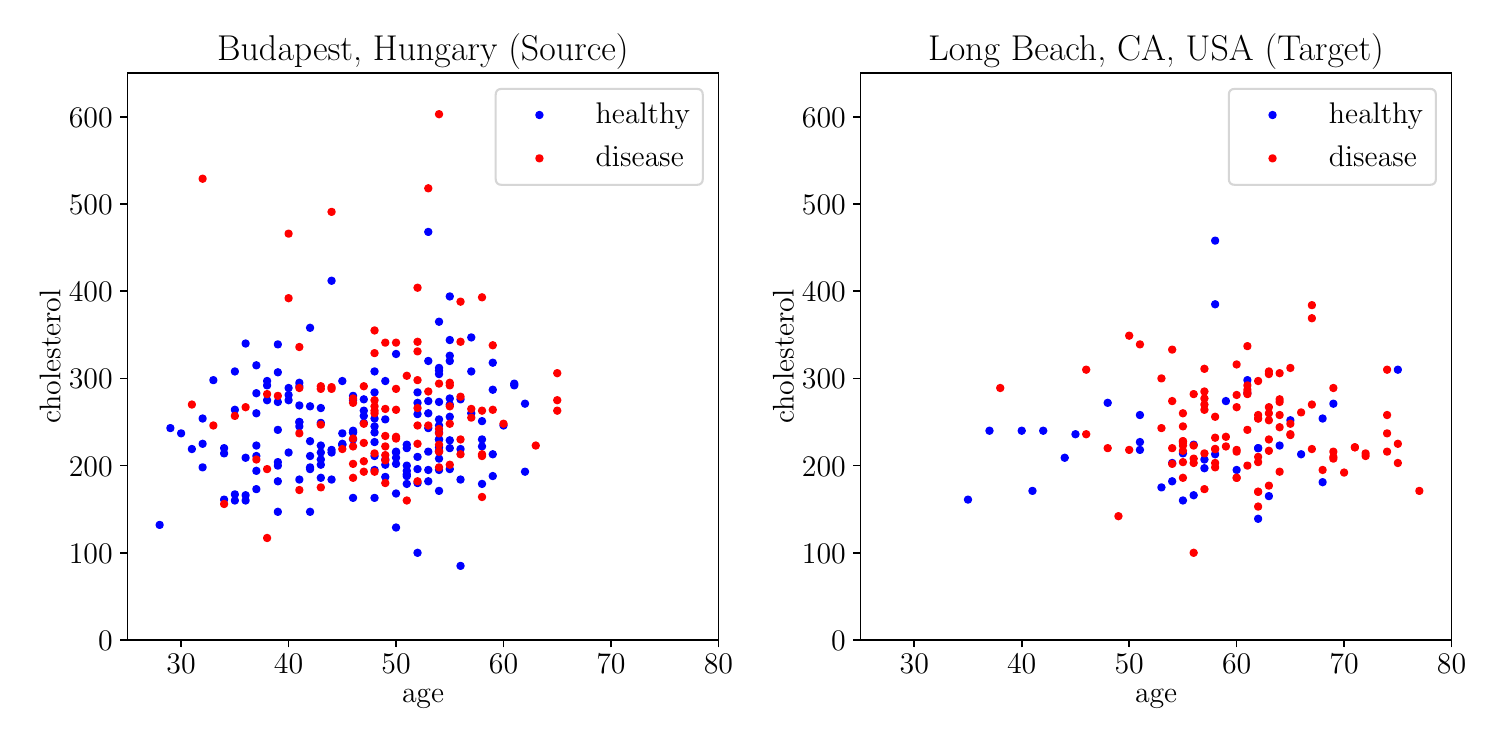}
    \caption{Actual values}
    \label{subfig:ActualValues}
    \end{subfigure}
    \begin{subfigure}[b]{0.75\textwidth}
    \centering
    \includegraphics[width=\textwidth]{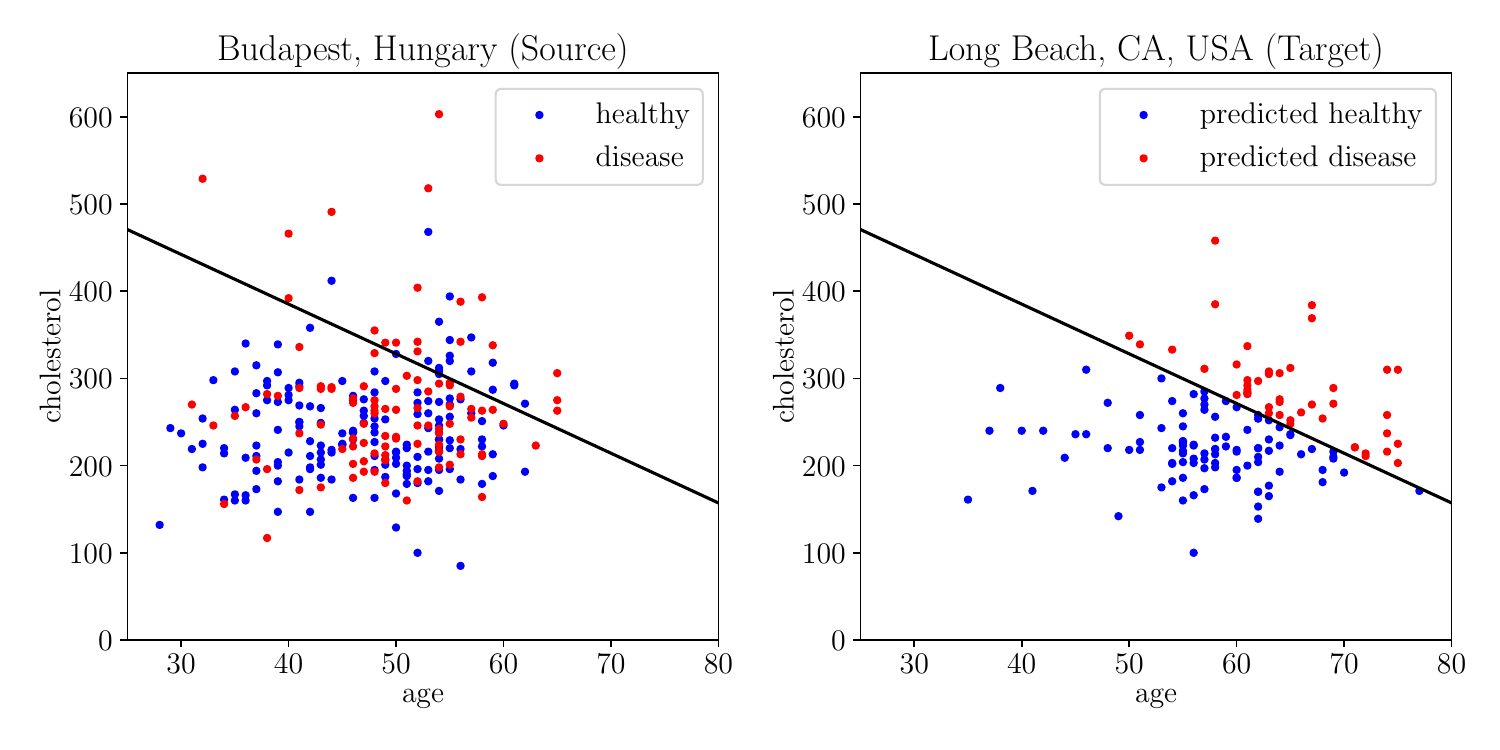}
    \caption{SVM trained on Budapest data}
    \label{subfig:SVMBudapest}
    \end{subfigure}
    \begin{subfigure}[b]{0.75\textwidth}
    \centering
    \includegraphics[width=\textwidth]{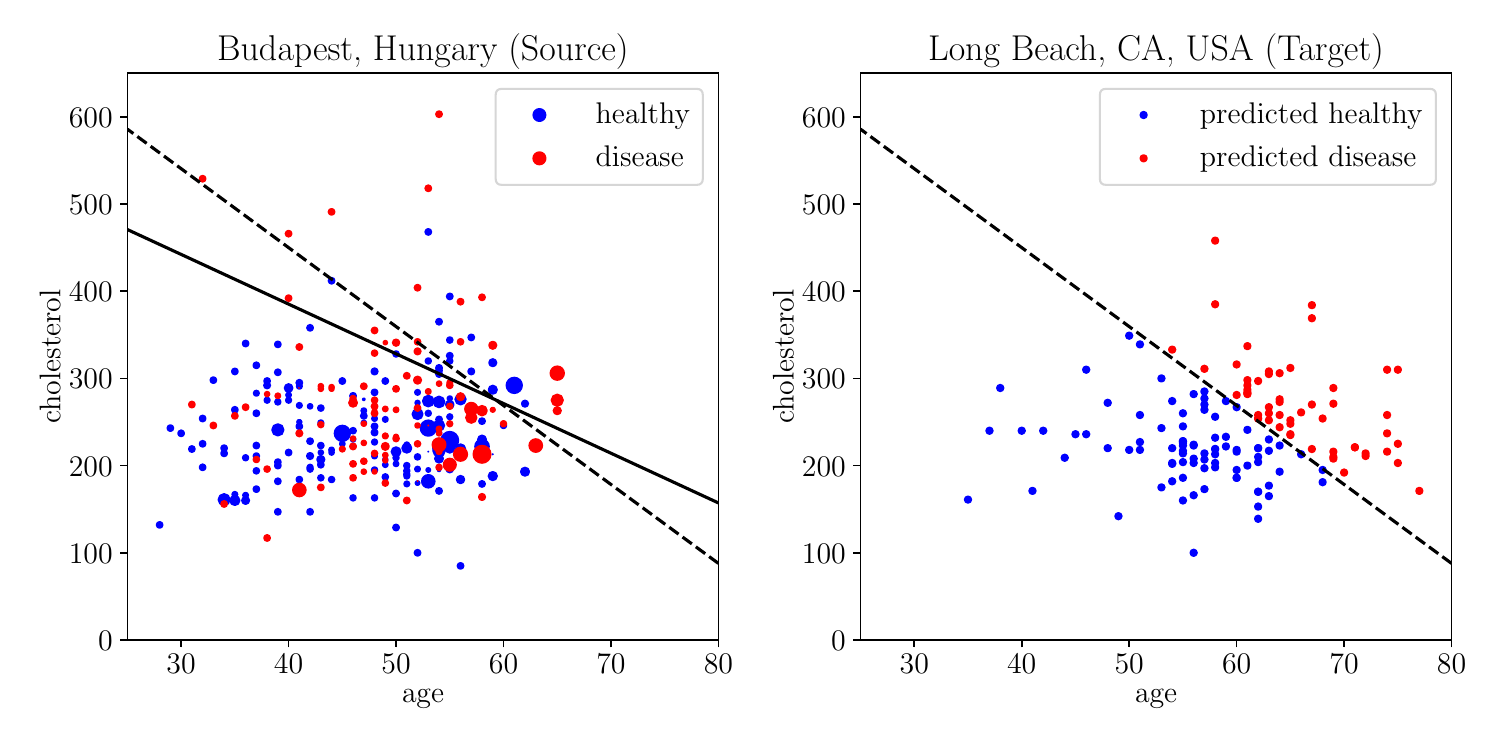}
    \caption{SVM trained on weighted Budapest data (dotted line)}
    \label{subfig:SVMWeightedBudapest}
    \end{subfigure}
    \caption[Heart disease prediction.]{The performance of a misspecified classifier can be improved by sample-based reweighting of the source to resemble the target distribution for a covariate shift more closely.}
    \label{fig:Cholesterol}
\end{figure*}

In a covariate shift, the right concept for both domains can be learned from only the source if the classifier is well-specified. \citet{kouw2019} use a linear classifier and observe worse than random performance in the target domain, which we were able to recreate. This means that the unadaptive classifier is not learning the right concept, either because it does not have the capacity, i.e., a linear decision boundary is not the right choice, or the attributes in the model do not have the power to explain the relationship entirely. It is unlikely that developing heart disease is only a matter of age and cholesterol level, nor is this relationship linear, so it is safe to assume that the classifier is misspecified. Statistical indicators of this are low overall accuracy and low generalization capabilities of the model. 
The frameworks suggest that the performance can be improved for misspecified classifiers by weighting the samples in the source domain to resemble the target domain. However, the authors do not publish the exact methods and results. In our experiments, for \ac{MMD} kernel matching \citep{borgwardt2006} using a \ac{RBF} kernel, the accuracy of the unweighted classifier is 0.47, and the weighted classifier achieves an accuracy of 0.54. \Cref{fig:Cholesterol} shows the decision boundaries and the learned weights. 

\subsubsection{Breast Cancer}
\label{sec:evaluation:naturalistic:breast_cancer}
In the Breast Cancer Wisconsin data set, image data of cell nuclei in women's breast tissue is described via 9 features, e.g., the symmetry or the radius of the nucleus. For each image, the label states if the cell belongs to a malignant cell, i.e., if it is cancer \citep{wolberg1995}. Applying the domain adaptation framework, we first analyze the causality. This problem represents a $Y \rightarrow X$ scenario, as the cancer is the cause for the transformation in the cells visible on the image data. In the example, all pictures are made under the same circumstances, and a prior shift is artificially introduced. This means the problem is no general data set shift or a class-conditional shift but a prior shift instance. 

\begin{table*}
\caption[Classification results of class-based reweighting approach.]{Classification results of \citet{saerens2002} class-based reweighting approach (\ac{EM}) versus a base-line reweighting via confusion matrix for detecting cancerous cells. The classifier in the source domain is a neural network that has been trained on a balanced data set.}
\label{tab:Saerens}
\centering
\small
\begin{tabular}{*{7}{c}}
\toprule
\multicolumn{1}{c}{\begin{tabular}[c]{@{}c@{}}True \\ priors\end{tabular}} &
  \multicolumn{2}{l}{Priors estimated by} &
  \multicolumn{4}{l}{Percentage of correct classification} \\ \cmidrule(l){2-3} \cmidrule(l){4-7} 
 & EM &
  \multicolumn{1}{c}{\begin{tabular}[c]{@{}c@{}}Confusion \\ matrix\end{tabular}} &
  \multicolumn{1}{c}{\begin{tabular}[c]{@{}c@{}}No \\ adjustment\end{tabular}} &
  \multicolumn{3}{c}{\begin{tabular}[c]{@{}c@{}}After adjustment\\ by using\end{tabular}} \\ \cmidrule(l){5-7} 
 & & & &EM &
  \multicolumn{1}{c}{\begin{tabular}[c]{@{}c@{}}Confusion \\ matrix\end{tabular}} &
  \multicolumn{1}{c}{\begin{tabular}[c]{@{}c@{}}True \\ priors\end{tabular}} \\ \midrule
20 & 18 & 26.2 & 91.3 & 92.0 & 92.1 & 92.6 \\ \bottomrule
\end{tabular}
\end{table*}

\citet{saerens2002} apply their \ac{EM} reweighting for a source domain with equal priors and a target domain with artificially changed posteriors and observe only a marginal improvement. \Cref{tab:Saerens} shows that the class-based reweighting has little effect, although it can more closely approximate the true priors than the alternative method. A possible explanation for this is that the overall accuracy of the neural network for both classes is already high, so the leverage of a class-based reweighting is small.

\subsubsection{MNIST, USPS, and SVHN Number Recognition}
\label{sec:evaluation:naturalistic:mnist}
In the following scenario, images of numbers from different sources must be correctly classified. Both the MNIST (Modified National Institute of Standards and Technology) and USPS data sets contain images of handwritten digits from 0 to 9. The images of MNIST are 28x28 gray-scale pixels, and the images of USPS are 16x16 gray-scale pixels \citep{hull1994}. Another data set that contains numbers is the Google Street View House Numbers (SVHN) data set. It consists of 32x32 pixel three-channel color images of digits, respectively numbers,  from 1 to 10 extracted from pictures of house numbers \citep{netzer2011}. In a preprocessing step not covered here, the image data must be transformed into the exact resolution and dimensionality so that the assumption of a heterogeneous domain adaptation problem, i.e., equal features, is fulfilled. For image data, the causality can only go from $Y \rightarrow X$ because the actual object (i.e., the written number) is the reason for the image data, not vice versa. 

In this artificial example, we know that for \emph{MNIST} and \emph{USPS}, all digits are present approximately equally in the data sets resulting in equal priors between both domains. In a real problem setting, this assumption might not be the case, e.g., if the digits in one domain originally belong to zip codes, leading zeros might not be allowed and thus resulting in a shift in the prior distribution. USPS is based on zip codes, but the classes are balanced. Therefore a reasonable assumption is a class-conditional shift. Intuitively, this makes sense because, in each domain, the images depict the same thing ($Y$). Still, the resulting image $X$ is different as a result of different handwriting or different camera parameters. Following the framework, domain adaptation can be accomplished by estimating the transformation function, which can be learned from aligning the distributions in the feature space. The success of the domain adaptation depends on two factors: (a) how close is the relationship between the domains, i.e., how stable is the transformation function, and (b) how well can a domain adaptation approach learn the actual transformation function. To answer (a), we can rely on expert knowledge: because humans can recognize numbers from both domains without problems, a stable transformation function exists. To answer (b) unfortunately is not possible a priori, because there is `no free lunch'\footnote{The `no free lunch' theorem introduced by \citet{wolpert1996} is a well-known principle in \ac{ML}, which essentially asserts that there is no universally superior learning algorithm. In other words, the effectiveness of a given ML algorithm is task-dependent: while one might perform excellently on certain problems, it may not fare as well on others. Thus, the phrase `no free lunch' signifies that there is no one-size-fits-all solution or shortcut in \ac{ML}---the optimal algorithm always depends on the specific problem at hand.} in \ac{ML} \citep{wolpert1996}. In their survey of unsupervised deep domain adaptation, \citet{wilson2020} collect empirical results for domain adaptation in both directions: MNIST to USPS and USPS to MNIST. \Cref{tab:NumbersComparison} shows an excerpt of this survey. In most cases, an accuracy close to the supervised case is achieved. There is no clear preference for a certain adaptation direction. 

\begin{figure}
\captionsetup[subfigure]{justification=centering}
    \centering
    \begin{subfigure}[b]{0.3\linewidth}
    \centering
    \includegraphics[width=\linewidth]{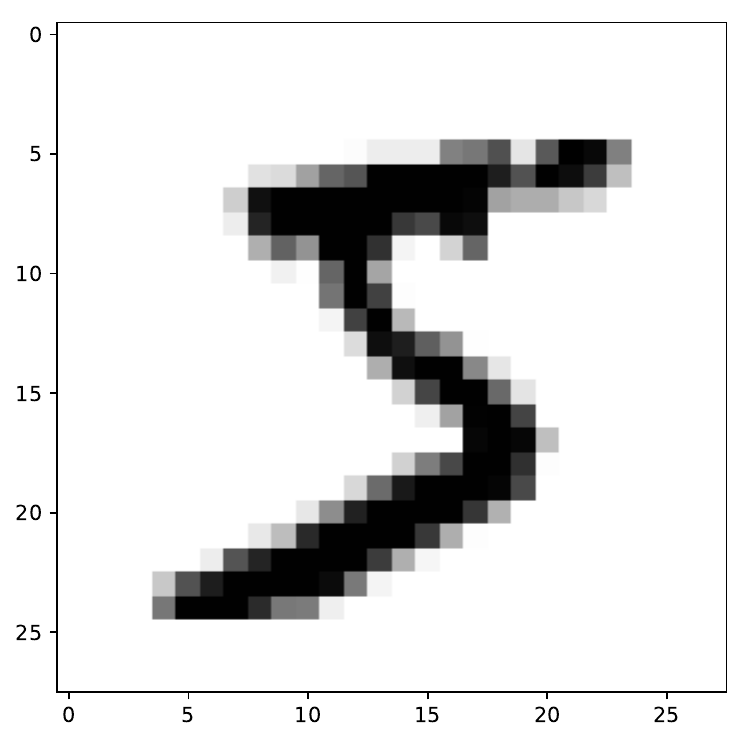}
    \caption{MNIST}
    \label{subfig:Mnist}
    \end{subfigure}
    \hfill
    \begin{subfigure}[b]{0.3\linewidth}
    \centering
    \includegraphics[width=\linewidth]{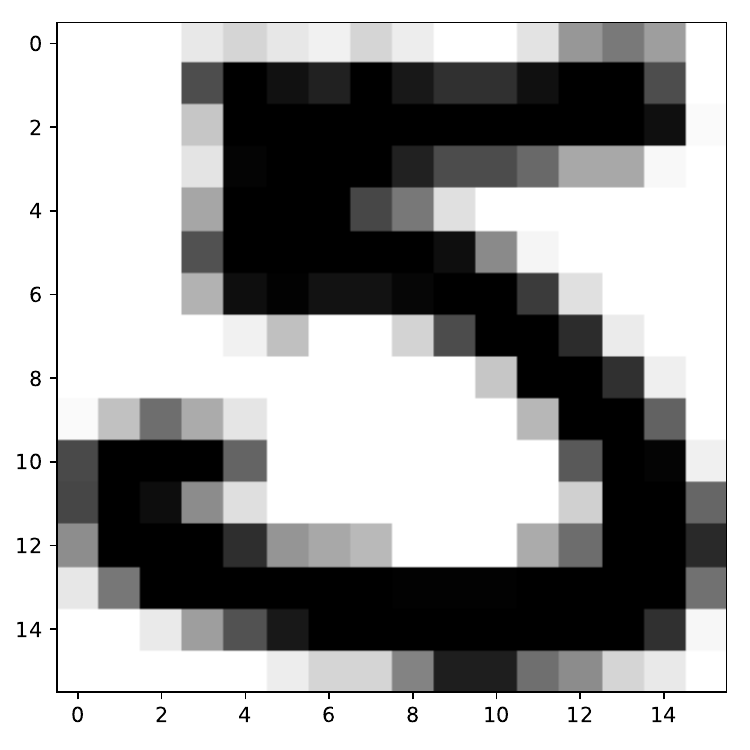}
    \caption{USPS}
    \label{subfig:Usps}
    \end{subfigure}
    \hfill
    \begin{subfigure}[b]{0.3\linewidth}
    \centering
    \includegraphics[width=\linewidth]{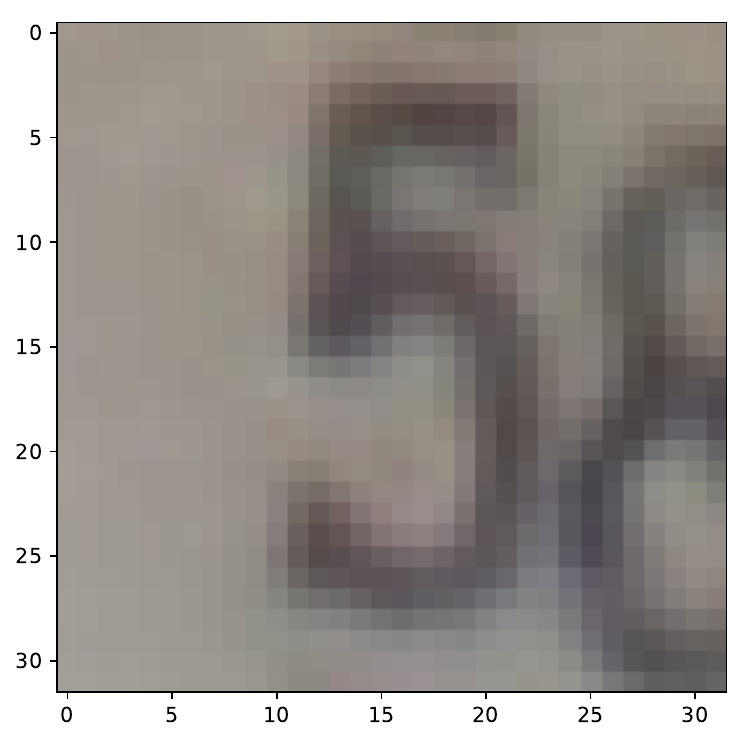}
    \caption{SVHN}
    \label{subfig:Svhn}
    \end{subfigure}
    \caption[Number recognition example.]{Example of class `5' from each data set.}
    \label{fig:Numbers}
\end{figure}

\begin{table*}
\caption[Classification accuracy for number recognition.]{Classification accuracy of different domain adaptation methods. Domains are presented in the form source $\rightarrow$ target. The survey was conducted by \citet{wilson2020}.}
\label{tab:NumbersComparison}
\centering
\begin{tabular}{l *{6}{c}}
\toprule
Name &
  \multicolumn{2}{c}{MNIST and USPS} &
  \multicolumn{2}{c}{MNIST and SVHN} \\ \cmidrule(lr){2-3} \cmidrule(lr){4-5}
 &
  MN $\rightarrow$ US &
  US $\rightarrow$ MN &
  SV $\rightarrow$ MN &
  MN $\rightarrow$ SV \\ \midrule
Target only \citep{hoffman2018}&
  \begin{tabular}[t]{@{}c@{}}96.3 \end{tabular} &
  99.2 &
  \begin{tabular}[t]{@{}c@{}}99.2\end{tabular} &
   \\ \midrule
\citet{french2018}& 98.2 & 99.5 & 99.3 &
  \begin{tabular}[t]{@{}c@{}}37.5\\ 97.0\end{tabular} \\
Co-DA \citep{kumar2018}& & & 98.6 & 81.7 \\
DIRT-T \citep{shu2018}& & & 99.4 & 76.5 \\
CyCADA \citep{hoffman2018}& 95.6 & 96.5 & 90.4 & \\
DRCN \citep{ghifary2016}& 91.8 & 73.7 & 82.0 & 40.1 \\ \midrule
Source only \citep{hoffman2018} &
  \begin{tabular}[t]{@{}c@{}}82.2\end{tabular} &
  69.6 &
  \begin{tabular}[t]{@{}c@{}}67.1\end{tabular} &
   \\ \bottomrule
\end{tabular}

\end{table*}

\begin{figure}
    \centering
    \includegraphics[width=0.7\linewidth]{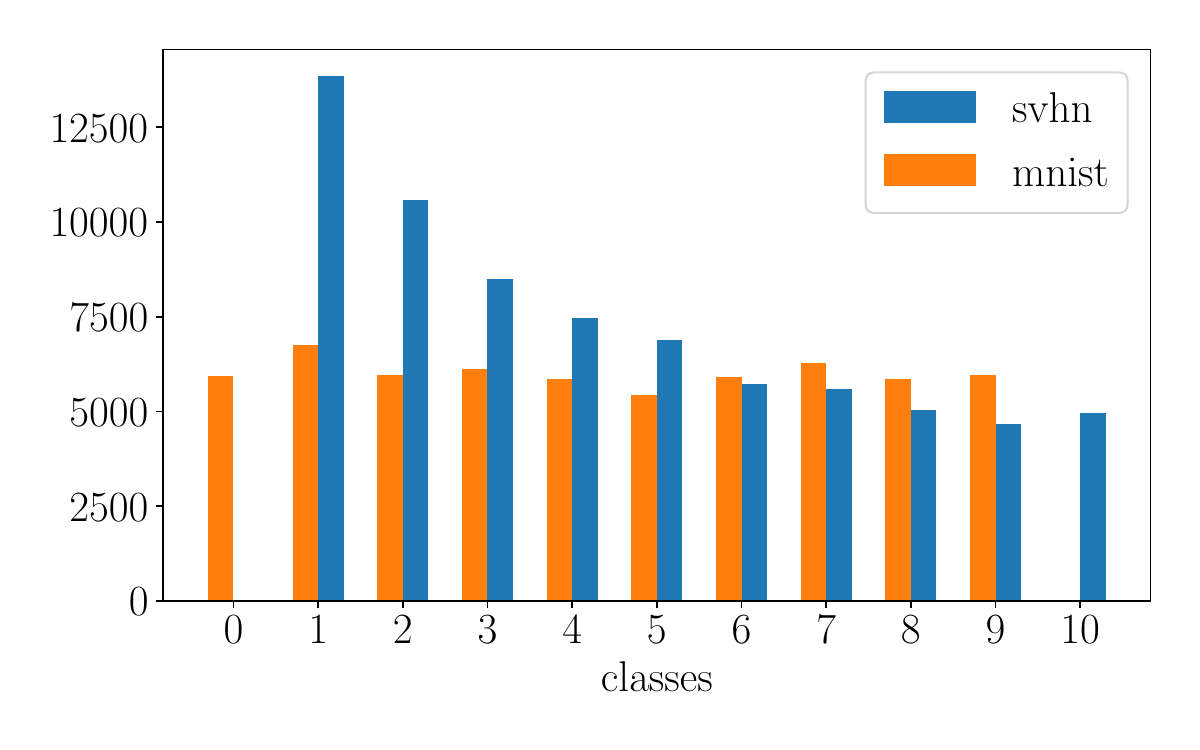}
    \caption[Prior distributions of SVHN and MNIST.]{The label distribution for MNIST and SVHN data set shows a class imbalance for SVHN. Because SVHN contains image data of house numbers, the leading digits cause an overrepresentation of lower numbers.}
    \label{fig:MnistSvhnImbalance}
\end{figure}

Going forward to the second domain adaptation case of \emph{MNIST and SVHN}, we observe that (a) the domain difference is intuitively greater as we compare handwritten gray-scale images to color photos of house numbers, and (b) the SVHN introduces a class-imbalance as shown in \Cref{fig:MnistSvhnImbalance}. Following the framework, this problem is a $Y \rightarrow X$ general data set shift scenario. This means while the solution procedures are the same as for the class-conditional shift, the quality of the solution is expected to be lower on average, depending on how well the class-conditional shift can be accounted for, how significant the prior shift is and if the model that is trained on the source domain makes systematically more mistakes for a particular class. Unfortunately, the reason cannot be pinned down without more profound research that is not available in the comparison provided by \citet{wilson2020}. The most striking difference in the comparison is the lower performance for the direction MNIST $\rightarrow$ SVHN for all domain adaptation approaches. The framework suggests that the drop-off in performance is caused either by the prior shift or by the fact that the transformation function is more challenging to learn in the MNIST $\rightarrow$ SVHN direction. The sensitivity to either of these parameters appears to depend on the approach. For example, \citet{kumar2018}, the authors of Co-DA, state difficulties in learning the concept, i.e., the transformation function. This is intuitively plausible, as the image data of SVHN is more complex and, therefore might be harder to learn if only image data of a simpler domain, e.g., MNIST, is available. On the other hand, \citet{french2018} state that their self-ensembling algorithm had problems adapting to the prior shift in SVHN.

\subsection{Evaluation Episode 3}
\label{sec:evaluation:ee3}
To assess the practical applicability and users' acceptance, we conduct an experimental study with a between-subject design. We measure the performance of users on exemplary domain adaptation tasks with and without decision support through our proposed framework. We are recruiting a total of 100 participants for our study in two rounds of 50 participants each---a first survey was conducted in July 2023 and a follow-up survey in January 2024---via the Prolific.co platform \citep{palan2018prolific}. Prolific is a widely acknowledged platform for acquiring reliable study participants for research endeavors \citep{peer2017beyond, palan2018prolific}. We split the participants randomly into two equally sized partitions (i.e., treatment groups).

\begin{figure*}
    \centering
    \includegraphics[width=\linewidth]{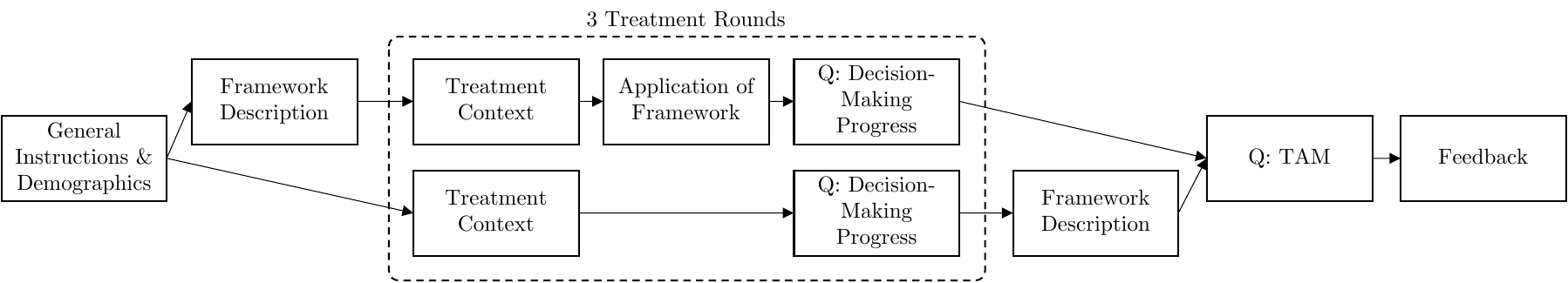}
    \caption[Between-subject design study.]{Between-subject evaluation study design. Two treatment groups with/without framework support assess framework applicability.}
    \label{fig:ee3-study}
\end{figure*}

As depicted in \Cref{fig:ee3-study}, the evaluation study followed a structured procedure involving two treatment groups. Prior to the main tasks, participants provided demographic information and received general instructions. The first treatment group was presented with an introductory overview of our domain adaptation framework, including explanations of causality and the different types of distribution shifts. They were then presented with a description of an exemplary domain adaptation case and asked to assess the causality and identify the relevant shift types based on the provided options. The system generated recommendations based on the selected options, and participants were required to select the appropriate domain shift scenario and corresponding method to overcome the drift, guided by the framework. The second treatment group received the same domain adaptation case description but did not receive any decision support from the framework initially. After three treatment rounds, presented in a randomized sequence, the second treatment group was also introduced to the framework. Subsequently, all participants were asked to evaluate their experience with the framework using the technology acceptance model (TAM) constructs: 'perceived usefulness' (U), `perceived ease of use' (E), `attitude towards using' (A), and `behavioral intention to use' (BI) measured with 10 items according to the suggestion of \citep{rigopoulos2008tam}. The assessment was conducted using a five-point Likert scale, and participants were given the opportunity to provide open-ended feedback through a free text form.

\Cref{tab:ee3-performance-comparison} presents a comparative analysis of the treatment groups with and without the framework. The study included 50 participants in the treatment group with the framework and 50 participants in the treatment group without the framework. The table illustrates the participants' performance in correctly identifying the domain shift in three evaluated cases: heart disease (covariate shift), spam detection (prior shift), and image recognition (covariate shift).

Within the treatment group supported by our proposed decision support framework, 28 participants demonstrated the ability to accurately identify the heart disease domain shift, while 37 participants correctly recognized the presence of the spam domain shift. Moreover, 21 participants displayed proficiency in identifying the domain shift related to image recognition. In contrast, the treatment group without the framework exhibited lower performance, with only 6 participants identifying the heart disease domain shift, 17 participants recognizing the spam domain shift, and 4 participants correctly identifying the image recognition domain shift. Thus, according to the Chi-squared independence test, the performances on all three tasks differs significantly ($p<0.01$) depending on whether the decision support framework was applied.

These findings highlight the positive influence of the framework on participants' capability to identify the relevant domain shifts. The higher number of correct responses in the treatment group with the framework signifies the framework's efficacy in providing valuable guidance and support in detecting and addressing domain shifts within the evaluated cases.

\begin{table}
\caption[Decision-making performance of study participants.]{The decision-making performance supported by our proposed framework is significantly higher compared to the treatment group without decision support.}
\label{tab:ee3-performance-comparison}
\centering
\small
\begin{tabular}{lrr}
\toprule
Treatment & w & w/o \\
\midrule
Heart Disease (Covariate Shift) & 28 & 6 \\
Spam Detection (Prior Shift) & 37 & 17 \\
Image Recognition (Class-cond. Shift) & 21 & 4 \\ \midrule
Responses overall & 50 & 50 \\
\bottomrule
\end{tabular}
\end{table}

\begin{figure}
    \centering
    \includegraphics[width=0.7\linewidth]{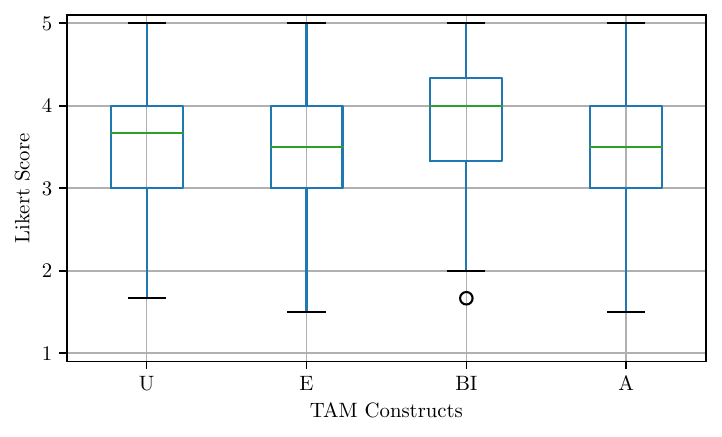}
    \caption[Survey responses for technology acceptance model constructs.]{Survey responses for technology acceptance model constructs.}
    \label{fig:ee3-tam-results}
\end{figure}

\Cref{fig:ee3-tam-results} supports the predominantly positive findings from the previous performance evaluation. Overall, the framework presented as an expert system is perceived as useful, as indicated by a mean score of 3.64 (sd = 0.78). Similarly, we observe positive feedback for the behavioral intention to use (mean = 3.80, sd = 0.75). In contrast, the construct of ease of use exhibits a higher standard deviation (0.93) and slightly lower mean value (3.47), which can be attributed to the survey-specific prototypical implementation. This is confirmed by a few comments in the free-text field. Users mentioned difficulties in understanding the concept of causality, with only little details presented, and further reported insufficient information about the cases presented. However, overall, also the attitude towards using the framework is generally positive (mean = 3.54, sd = 0.81).

\section{Discussion}
\label{sec:discussion}

The research questions \emph{``How can the appropriateness of domain adaptation for a specific problem be determined a priori using practical indicators?} (RQ \ref{rq:1}) and \emph{``How can domain adaptation problems be meaningfully distinguished to suggest targeted solution strategies?''} (RQ \ref{rq:2}) can be approached by leveraging the presented framework and its evaluation insights. 

Starting with RQ \ref{rq:2}, five domain adaptation scenarios are identified based on the causal properties of the real system and the domain relationship, i.e., data shift (cf. \Cref{tab:scenarios}). Different solution procedures for each scenario are identified (cf. \Cref{tab:SolutionProcedures}) by literature review and motivated further by acknowledging the underlying theory, e.g., bounds for the generalization error.  During the first \ac{EE}, the theoretical implications are tested, and in the second \ac{EE}, the application of the framework demonstrates that it possesses explanatory capabilities. The evaluation also indicates that in some domain adaptation scenarios, choosing the wrong approach yields worse results, e.g., choosing a covariate shift approach if, in reality, a class-conditional shift occurred.  Even if the theory suggests this is the case for all scenarios, this claim is yet to be verified in future contributions. In the third \ac{EE}, although no full-scale implementation of a realistic expert system has been evaluated, we demonstrate that following the framework suggestions delivers significant benefit over an uninformed domain adaptation attempt. As for solution recommendations, it becomes apparent that the expressiveness depends on the complexity of the domain adaptation problem. For simpler shifts, like covariate and prior, there even exist some statistical tests that indicate if a certain solution approach is beneficial. In the case of more complex scenarios, like class-conditional or general shift, the approximation of a transformation function is non-trivial, and there is no free lunch \citep{wolpert1996, kouw2019}.
In conclusion: \emph{Yes, it is possible to differentiate between domain adaptation problems meaningfully. It is also possible to recommend suitable (family or) families of solution approaches for each scenario. However, for complex scenarios, precise implementations cannot be recommended because their performance depends too much on certain technicalities, i.e., there is no free lunch for domain adaptation.}

Answering RQ \ref{rq:1} requires analyzing how well the five scenarios can be identified in a practical \ac{ML} scenario under common restrictions, e.g., no labels in the target domain. \Cref{tab:ScenarioDetermination} summarizes a set of approaches to identify each scenario. To do so, one must first define the causality of the system, which leads to the first instance of expert knowledge that is required. Further, instead of relying on theoretical divergence measures or their practical implementations that exhibit different forms of limitations depending on which approach is used, the framework relies on expert knowledge about the real system to give an educated estimation of probable relationships, e.g., if a concept shift is possible. This is the second instance of expert knowledge. After establishing a frame of plausible relationships between the domains, some of these assumptions, e.g., a covariate shift, can be tested a priori, while others---e.g., a class-conditional shift---cannot. Even a posteriori, the failure of a domain adaptation approach does not always prove that domain adaptation is not possible. Instead, the chosen implementation might have failed because, e.g., in a class-conditional problem, the transformation function could not be sufficiently learned. But in all cases, the framework provides a theoretical grounding on which success or failure can be interpreted more productively and consecutively, leading to deeper insights into the problem at hand, which is a benefit over uninformed domain adaptation. In conclusion, we could not find a way to determine the suitability of domain adaptation for any given problem (by indicators that are available a priori) because of two reasons: (1) The identification of scenarios relies on expert knowledge that cannot always be statistically verified, and (2) the failure of a domain adaptation attempt does not mean that other attempts will also fail. But even for negative results, the framework allows practitioners to narrow down possible reasons for failure, increasing the probability of future success.

Thus, the presented research offers a practical problem-oriented domain adaptation framework and showcases its potential to assist \ac{ML} practitioners. However, this research also has limitations.

Firstly, as detailed above, one crucial limitation lies in the framework's heavy reliance on expert knowledge in identifying domain adaptation scenarios and their probable relationships. Such knowledge cannot always be statistically verified, and its availability may vary greatly across different practical contexts. Therefore, the framework may not be universally applicable, or its effectiveness may be compromised without adequate expertise.

Secondly, while the framework manages to recommend solution approaches for different scenarios for rather complex shifts, such as class-conditional or general shifts, it fails to provide precise implementation recommendations. As per the `no free lunch' theorem, the performance of specific implementations is highly contingent on the unique technicalities of each problem. This necessitates further research to devise methodologies for recommending precise implementations, particularly for complex scenarios.

Finally, the third evaluation episode involved a relatively small number of participants, which may limit the generalizability of the findings. Additionally, the evaluation was based on a prototype implementation of the framework, and a real-world implementation would likely involve a more comprehensive and user-friendly tool, such as a web-based platform or integration into existing workflows. It is important to note that while the framework has undergone artificial and naturalistic evaluation episodes, a larger-scale summative evaluation is still needed to empirically validate the assumption that adhering to the framework would significantly enhance the effectiveness of domain adaptation efforts compared to uninformed attempts. Future research should focus on conducting more extensive evaluations involving a broader range of participants and real-world implementation scenarios to further validate and refine the framework's utility and impact.

Given these limitations, the framework does offer significant contributions to the field of \ac{ML}, providing a structured perspective on domain adaptation problems and a unified overview of solutions to overcome them. In future work, these limitations can be addressed by incorporating a user-centric view in the evaluation, empirically testing more theoretical insights, and understanding the success or failure of different domain adaptation approaches in specific applications.

\section{Conclusion}
\label{sec:conclusion}
Guided by \ac{DSR}, we created a problem-oriented domain adaptation framework to answer the questions on how to determine the \emph{appropriateness} of domain adaptation a priori and consecutively give tailored \emph{solution recommendations} for \ac{ML} practitioners. To this end, the framework identifies five domain adaptation scenarios (cf. \Cref{tab:scenarios}) that are based upon the unified view of data set shifts by \citet{moreno-torres2012}: prior shift, class-conditional shift, covariate shift, concept shift, and general shift.

For each scenario, one or multiple solution approaches (cf. \Cref{tab:SolutionProcedures})  can be recommended based on the scenario properties. For complex problems, the performance can vary heavily depending on the implementation details, so it is not possible to estimate the result for any domain adaptation implementation a priori. For the same reason, the failure of one domain adaptation attempt does not mean this will be the case for all domain adaptation approaches. Further, the determination of scenarios (cf. \Cref{tab:ScenarioDetermination}) relies on understanding the causality of the system and, as statistical indicators are prone to error, expert knowledge about the domains can be hard to verify objectively.  

In summation, while it remains challenging to determine a problem's suitability for domain adaptation, particularly given the obstacles of objective scenario determination and the inherent complexity of diverse solution procedures, the framework offers three substantial contributions.

First, it provides a structured perspective on domain adaptation, which is of considerable benefit to practitioners struggling with the complexity of the field. We demonstrate the efficacy of the proposed framework in three \acp{EE}, verifying the framework's capacity to support practitioners. Second, it provides a selection of suitable solution approaches categorized by problem types, serving as a practical guide in navigating potential strategies. Third, the framework serves as a decision support tool in real-world applications, providing crucial direction and strategy suggestions for tackling domain adaptation problems. This makes it an improvement over the status quo of uninformed domain adaptation.

From a \ac{CS} perspective, our framework contributes to refining domain adaptation theory. Utilizing an established perspective on dataset shifts provides a means to differentiate various problem settings effectively and thus facilitates the enhancement of theoretical understanding of domain adaptation.

Future work can enhance the framework's evaluation by adding a user-centric perspective, which may involve assessing the user's perceived benefits through surveys or workshop experiments. There remains a wealth of untested theoretical insights that need empirical validation, such as quantifying the advantage of correct domain adaptation selection. Furthermore, an in-depth exploration of problem characteristics within the domain adaptation framework, the performance evaluation of cutting-edge domain adaptation implementations, and a better understanding of why and how specific domain adaptation approaches succeed or fail are all promising future research directions. This could facilitate the refinement of domain adaptation procedures, especially in class-conditional or general dataset shifts.

\appendix

\section*{Acknowledgements}

Generative AI tools were utilized throughout this work. Specifically, DeepL Write, and Grammarly were used to enhance the writing quality of tutorials and explanations provided to participants during the experiments, as well as to improve the language across all sections of this paper.

\bibliographystyle{elsarticle-harv} 
\bibliography{mylibrary}

\end{document}